\documentclass[11pt]{article}

\usepackage{arxiv}
\usepackage[utf8]{inputenc}
\usepackage[T1]{fontenc}
\usepackage[numbers,sort&compress]{natbib}
\usepackage{graphicx}
\usepackage{amsmath}
\usepackage[defaultsups=false]{newtxtext}
\usepackage{newtxmath}
\makeatletter
\DeclareRobustCommand*\textsuperscript[1]{\m@th\ensuremath{^{\mbox{\fontsize\sf@size\z@\selectfont#1}}}}
\makeatother

\makeatletter
\renewcommand{\@toptitlebar}{%
  \hrule height 4\p@
  \vskip 0.25in \vskip -\parskip}
\renewcommand{\@bottomtitlebar}{%
  \vskip 0.29in \vskip -\parskip
  \hrule height 1\p@
  \vskip 0.09in}
\renewcommand{\@maketitle}{%
  \vbox{%
    \hsize\textwidth
    \linewidth\hsize
    \vskip 0.1in
    \@toptitlebar
    \centering
    {\LARGE\bfseries \@title\par}
    \@bottomtitlebar
    \vskip 0.06in
    \def\And{\end{tabular}\hfil\linebreak[0]\hfil\begin{tabular}[t]{c}\bf\rule{\z@}{24\p@}\ignorespaces}
    \def\AND{\end{tabular}\hfil\linebreak[4]\hfil\begin{tabular}[t]{c}\bf\rule{\z@}{24\p@}\ignorespaces}
    \begin{tabular}[t]{c}\bf\rule{\z@}{24\p@}\@author\end{tabular}%
    \vskip 0.28in \@minus 0.1in
  }
}
\makeatother
\usepackage{array}
\usepackage{multirow}
\usepackage{makecell}
\usepackage{booktabs}
\usepackage{float}
\usepackage{needspace}
\usepackage{xcolor}
\definecolor{ink}{HTML}{16283F}
\definecolor{accent}{HTML}{15588C}
\definecolor{blockbg}{HTML}{F5F6F8}
\usepackage[ruled,vlined]{algorithm2e}
\usepackage{listings}
\usepackage{microtype}
\usepackage{url}
\usepackage{hyperref}
\usepackage{titlesec}
\titleformat{\section}{\color{ink}\Large\bfseries}{\thesection}{0.85em}{}
\titleformat{\subsection}{\color{ink}\large\bfseries}{\thesubsection}{0.85em}{}
\titleformat{\subsubsection}{\color{ink}\normalsize\bfseries}{\thesubsubsection}{0.85em}{}
\titlespacing*{\section}{0pt}{2.4ex plus .6ex minus .2ex}{1.3ex plus .3ex}
\titlespacing*{\subsection}{0pt}{2.0ex plus .5ex minus .2ex}{0.9ex plus .2ex}
\titlespacing*{\subsubsection}{0pt}{1.7ex plus .4ex minus .2ex}{0.6ex plus .2ex}
\usepackage[font=small,labelfont={bf,color=ink},skip=8pt]{caption}

\SetAlCapFnt{\color{ink}\small\bfseries}
\SetAlCapNameFnt{\small}

\title{Verifier-Based Reinforcement Fine-Tuning of Reasoning Models for Thermal Energy Storage Control}

\renewcommand{\headeright}{}

\makeatletter
\def\keywordname{{\color{ink}\bfseries\scshape Keywords}}
\def\keywords#1{\par\addvspace\medskipamount{\small\rightskip=0pt plus1cm
\def\and{\ifhmode\unskip\nobreak\fi\ {\color{accent}\textbullet}\ }%
\noindent\keywordname\quad\ignorespaces#1\par}}
\makeatother
\renewcommand{\undertitle}{}
\renewcommand{\shorttitle}{Verifier-Based RFT for TES Control}

\author{%
  Takumi Shioda\thanks{Corresponding author. ORCID: \href{https://orcid.org/0009-0007-7335-4394}{0009-0007-7335-4394}.} \\
  The University of Tokyo \\
  \texttt{tkm-0211@iis.u-tokyo.ac.jp} \\
  \And
  Kohei Terashima \\
  Tokyo University of Science \\
  \texttt{k.terashima@rs.tus.ac.jp} \\
  \And
  Tatsuo Nagai \\
  Tokyo University of Science \\
  \texttt{nagai@rs.tus.ac.jp} \\
}

\date{}

\hypersetup{
  colorlinks=true,
  linkcolor=accent,
  citecolor=accent,
  urlcolor=accent,
  pdftitle={Verifier-Based Reinforcement Fine-Tuning of Reasoning Models for Thermal Energy Storage Control},
  pdfauthor={Takumi Shioda, Kohei Terashima, Tatsuo Nagai},
  pdfkeywords={reinforcement learning with verifiable rewards, reinforcement fine-tuning, reasoning models, large language models, thermal energy storage, dynamic programming, energy management}
}

\lstdefinestyle{promptblock}{
  basicstyle=\ttfamily\footnotesize,
  breaklines=true,
  breakatwhitespace=false,
  columns=fullflexible,
  keepspaces=true,
  showstringspaces=false,
  backgroundcolor=\color{blockbg},
  frame=leftline,
  framerule=1.5pt,
  rulecolor=\color{accent},
  xleftmargin=12pt,
  framexleftmargin=8pt,
  framextopmargin=4pt,
  framexbottommargin=4pt,
  aboveskip=9pt,
  belowskip=9pt
}

\newcommand{\promptrole}[1]{\par\Needspace*{6\baselineskip}\noindent\textbf{#1}\par\nopagebreak}

\begin{document}
\maketitle

\begin{abstract}
Buildings are expected to shift cooling loads in response to grid conditions. Thermal energy storage (TES) enables this shift, but scheduling it well requires planning hours ahead under storage constraints. Model predictive control (MPC) and reinforcement learning are difficult to scale across buildings. This study instead adapts an open-weight reasoning model through reinforcement learning with verifiable rewards (RLVR). We convert exact offline dynamic-programming (DP) action values into dense rewards for every candidate action. Using only 30 training prompts, reinforcement fine-tuning (RFT) trains the model as an upper-level scheduler that outputs hourly heat-pump setpoints from text-based states and forecasts. Evaluation uses a deliberately simple office-building TES benchmark where exact DP is tractable and the optimum is known. RFT reduces the open-weight model's emissions from 70.5 to 61.2 kg-CO$_2$, close to the DP optimum of 60.8 kg-CO$_2$. GPT-5 nearly matches DP and MPC without task-specific training, while GPT-4o, a non-reasoning LLM, produces higher emissions than the no-storage baseline, so inference-time reasoning appears important. Trace analysis shows that RFT mainly stabilizes observable planning patterns (candidate comparison, look-ahead, and feasibility checking) rather than creating a new strategy. Robustness and generalization tests clarify what transfers: the reinforced planning patterns persist under forecast errors and an unseen TES condition and carry over to a battery task, but its different structure limits the gains. DP-based verifiable rewards offer a practical way to adapt open-weight reasoning models to building storage scheduling. These results motivate higher-fidelity tests of whole-building control and scalable verifiers for city-scale energy management.
\end{abstract}

\keywords{Reinforcement learning with verifiable rewards \and Reinforcement fine-tuning \and Reasoning models \and Large language models \and Thermal energy storage \and Dynamic programming \and Energy management}

\section{Introduction}\label{sec:introduction}
Buildings are increasingly expected to provide not only energy efficiency but also operational flexibility \cite{ref:ipcc_ar6_wg3}. With variable renewable generation expanding, building demand response plays a growing role in balancing supply and demand \cite{ref:jurjevic2023_dr_buildings}. This trend has motivated systematic methods for managing flexibility resources in buildings \cite{ref:pedram2023_flex_energies}. Building operations account for a substantial share of global final energy consumption \cite{ref:iea_buildings_energy_system}. Heating, ventilation, and air-conditioning (HVAC) systems are therefore central to reducing and reshaping that demand. From a grid-interactive building perspective, their controllable loads can be coordinated through scheduling and storage \cite{ref:bayasgalan2024_gieb_energies}. Thermal energy storage (TES) is particularly attractive for shifting cooling production from peak to off-peak periods \cite{sun2013_ctes_review}. When the scheduling signal is time-varying carbon intensity, the same storage mechanism can reduce operational emissions by shifting heat-pump operation to lower-carbon hours \cite{miyata2020_dynamic_co2_mpc}. Effective TES scheduling, however, is a sequential decision problem that requires the controller to compare candidate actions over time while respecting equipment and storage constraints.

Conventional optimization and learning approaches each have distinct strengths, but both face practical barriers. Model predictive control (MPC) offers principled look-ahead optimization and explicit constraint handling \cite{ref:serale2018_mpc_energies}. Its effectiveness in buildings nevertheless depends on an accurate system model and repeated online optimization \cite{ref:drgona2020_mpc_buildings}, while building and validating such models imposes nontrivial data requirements \cite{ref:zhan2021_mpc_data_requirements}. Reinforcement learning (RL) can in principle learn strong building-control policies from interaction data, but training cost, safety, and generalization remain practical challenges \cite{ref:wang2020_rl_building_controls}. Evidence from HVAC applications similarly shows that deployment requires careful choices of state representation, reward design, and training environment \cite{ref:sierla2022_rl_hvac_energies}. These requirements can turn deployment into a multi-stage engineering process involving data preparation, model development, and validation \cite{ref:drgona2020_mpc_buildings}. Field implementation adds commissioning, integration, and post-deployment retuning challenges that make predictive control difficult to scale across buildings \cite{ref:saloux2025_mpc_field_implementations}. Although many buildings now collect substantial operational data through building energy management systems (BEMS), such data are used mainly for monitoring and diagnostics rather than for predictive scheduling \cite{ref:savadkoohi2024_predictive_control_data}, leaving the loop from collected data to control decisions open. These limitations motivate controllers that can close this loop by acting on readily available operational information and that, once trained offline, could transfer across buildings---removing the need for site-specific modeling, repeated online optimization, or large-scale retraining at each new site.
\subsection{Background and motivation}\label{subsec:background}
Recent large language models (LLMs) have been developed to perform extended multi-step deliberation at inference time; we refer to such models as \textit{reasoning models}. OpenAI's introduction of o1 explicitly emphasized learning to reason through additional inference-time computation \cite{ref:openai2024_learning_to_reason}, while its system card documents the capabilities and limitations of that reasoning-oriented model family \cite{ref:openai2024_o1_system_card}. DeepSeek-R1 subsequently demonstrated that reinforcement learning can elicit and strengthen reasoning behavior in an open model \cite{ref:deepseekr1_2025}. This shift motivates the question of whether a reasoning model can serve as a controller for planning problems in which the decision maker must process structured state information, anticipate future consequences, and choose actions under operational constraints.

If reasoning models are to serve as such controllers, the next question is whether their decision behavior can be improved through task-grounded reinforcement fine-tuning (RFT). RFT is especially well suited to domains where candidate outputs can be checked against objective, automatically verifiable criteria rather than human preference. In game playing, explicit wins and losses provide unambiguous reinforcement signals \cite{ref:silver2016_alphago_nature}. Mathematical reasoning offers similarly checkable feedback through step-level verification \cite{ref:lightman2023_verify_step} and automated process supervision \cite{ref:luo2024_automated_process_supervision}. Group-relative policy optimization was introduced with automatically checkable mathematical feedback \cite{deepseekmath2024}. Subsequent work shows that verifiable rewards can improve the correctness of reasoning in base LLMs \cite{ref:wen2025_rlvr_correct_reasoning}, while theoretical analysis characterizes how group-relative updates amplify successful samples \cite{mroueh2025rlvr}. The key commonality across these cases is not the application domain itself, but the presence of a verifier that can assign reliable reward signals to candidate outputs. Energy-control decisions share this structure because they are evaluated against explicit physical constraints and long-horizon objectives, so reinforcement learning with verifiable rewards (RLVR) should in principle extend to energy control. In storage scheduling, a model-based optimizer can score every candidate action, not just the optimal trajectory, turning the control objective into a dense verifier for post-training. Related evidence shows that RL-based post-training can generalize where supervised fine-tuning memorizes \cite{ref:chu2025_sft_rl}, while carefully curated examples can elicit strong reasoning from limited data \cite{ref:ye2025_limo}. These findings do not establish transfer to control, but they motivate testing whether a small set of verified control states can reinforce a recurring planning procedure rather than isolated action labels.

TES scheduling provides a controlled testbed for this idea. Effective operation requires the controller to decide when to charge and discharge storage by jointly considering the current storage level, future cooling demand, time-varying carbon intensity, and capacity limits. The task is simple enough to keep action consequences transparent, yet rich enough that greedy short-horizon decisions become suboptimal. With a transparent energy-balance model and a compact discrete action space, exact backward dynamic programming (DP) remains tractable and can score every candidate action at every reachable state. We use this deliberately tractable benchmark because its action values can serve both as training rewards and as the optimal reference for evaluation. Extending the framework to higher-fidelity TES and whole-building systems is the next step; the corresponding verifier requirements and scope boundaries are discussed in Sections~\ref{subsec:framework-outlook} and~\ref{subsec:limitations}. The purpose is not to replace DP on a problem it already solves, but to test whether its offline action-value signal can reinforce reusable planning behavior in a reasoning model.

\subsection{Research questions and objectives}\label{subsec:gap-objectives}
LLMs have already been explored across energy systems: a recent review catalogs a broad range of applications \cite{ref:mirshekali2025_llm_energy_systems_review}, and individual studies apply LLMs to energy forecasting \cite{ref:qiu2024_ef_llm}, agent-based energy-system balancing \cite{ref:ren2025_llm_balance_energy_systems}, and translation of natural-language requirements into validated power-system optimization models \cite{ref:hu2025_solver_ready_power_system_optimization}. Closer to building operation, pretrained language models have been proposed for industrial control-policy generation \cite{ref:song2023_pretrained_llm_industrial_control}, and ChatGPT-based approaches have been evaluated for autonomous HVAC operation \cite{ref:ahn2023_chatgpt_hvac}. Decision-pretrained transformers offer a related route to HVAC control \cite{ref:berkes2024_hvac_dpt}, domain-guided retrieval has been introduced to make LLM-based HVAC decisions more interpretable \cite{ref:kojain2025_darlin}, and multimodal foundation models have been tested in office-in-the-loop HVAC control \cite{ref:sawada2024_office_in_the_loop} and, subsequently, agentic HVAC operation \cite{ref:sawada2025_agentic_ai_dce}. Other studies address automatic generation of EnergyPlus models \cite{ref:jiang2024_eplus_llm} or assess LLM knowledge in the HVAC domain \cite{ref:lu2024_ashrae_exam_llm}. Two gaps remain across these strands. First, most of these systems rely on direct response generation or instruction following, without allocating inference-time computation to extended multi-step planning. Second, to our knowledge, none derives a verifiable reward from the energy-control problem itself and uses it to fine-tune the model. Whether RLVR, which has proven effective where automatic verification is available, can be instantiated for constrained energy-management planning remains untested, and it is likewise unclear whether the multi-step reasoning capability of recent models is necessary for such planning at all.

Given these gaps, this study asks one central question: can the objective of a constrained storage-scheduling problem be converted into a verifiable reward, and does RFT with that reward bring an open-weight reasoning model close to exact optimization? We answer this by comparing controllers that all share the same physical model, action space, and objective. Our working hypothesis is that verifier-based RFT can teach the model a general way of approaching the scheduling problem rather than case-specific answers, so that effective adaptation should be possible from a small number of training prompts and should partially generalize to unseen operating conditions. We expect transfer to be strongest when the target setting preserves the decision structure represented in the verifier.

The primary subject of this study is an open-weight reasoning model evaluated before and after verifier-based RFT, with GPT-5 as a frontier reference for what is attainable without task-specific training and GPT-4o as a non-reasoning baseline that checks whether multi-step reasoning is needed for this task. The open-weight focus also matches deployment needs: data can stay local, inference is cheaper, and reasoning traces can be inspected.

The contributions are (i) a DP-based verifier that converts exact action values into dense, graded rewards for every admissible action at each verified state; (ii) a controlled evaluation of the open-weight model before and after RFT against DP, MPC, DQN, a frontier reasoning model, and a non-reasoning LLM under the same conditions; and (iii) a robustness and transfer analysis that tracks performance and observable planning patterns across forecast error, a shifted TES numerical regime, and a battery task with different physical semantics.
\section{Methodology}\label{sec:methodology}
This section describes the two experimental components used to address the central questions in Section~\ref{subsec:gap-objectives}: the shared LLM controller interface through which every language model in the comparison acts as a constrained storage scheduler (Section~\ref{subsec:controller-design}), and the verifier-based RFT framework and training procedure for the open-weight model (Sections~\ref{subsec:rft-procedure} and \ref{subsec:rft-training-procedure}).

\begin{figure}[t]
\centering
\includegraphics[width=\textwidth]{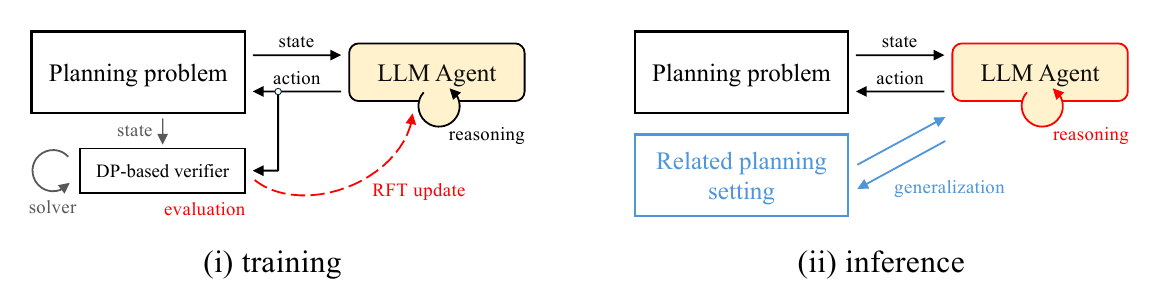}
\caption{Overview of the proposed verifier-based post-training framework, combining an offline DP verifier at training time with a post-trained LLM controller at inference time.}
\label{fig:planning_rft_framework}
\end{figure}

\subsection{LLM controller design}\label{subsec:controller-design}
We use LLMs, and in particular reasoning models, as controllers that select one action at each time step from the same discrete action set as the optimization and learning baselines. To support consistent comparison with these baselines, including MPC and RL, the information given to the LLM is aligned with the observation design used in the simulation environment: the current TES state, time information, and the forecast-related decision context needed for constrained planning. The observation is converted into structured text rather than passed as raw numerical arrays, because the goal is to test whether the model can read heterogeneous control information and use it for sequential decision making. In deployment terms, the LLM is positioned as an upper-level scheduler above the BEMS, not as an actuator-level controller: it issues an hourly ASHP output-ratio setpoint, and translating that setpoint into continuous equipment commands is delegated to conventional lower-level loops such as PID or DDC controllers.

The prompt is structured into three roles. A stable system message defines the model as a TES optimization agent. A developer-level instruction constrains the output format so that the model must emit one valid action in a machine-readable form. For the open-weight model, the executed final output is a single action token, while the reasoning text generated in the model's separate reasoning channel is retained for later analysis; for GPT-5, the output is a one-line JSON object containing a brief reasoning summary and the action. The user message then provides the current state, forecasts, and task constraints. This separation keeps the task definition stable while the state description changes at each step, reduces parsing ambiguity, and supports consistent control across repeated calls.

Only the parsed action affects the environment. After the model returns an action token or JSON object, the simulator extracts the action, applies the shared transition model, and advances to the next state. Representative prompt templates, output constraints, and forecast-error warning text are provided in Appendix~\ref{app:prompts}.
\subsection{Verifier-based RFT framework}\label{subsec:rft-procedure}
RFT is applied to the open-weight model by viewing the LLM as a language policy $\pi_{\theta}(y \mid x)$, where $x$ is the text-based TES state and $y$ is the generated response from which the discrete control action is parsed. The key requirement for RLVR-style post-training is a verifier that can assign deterministic rewards to candidate outputs (Section~\ref{subsec:background}). Here, that verifier is constructed directly from the TES control problem. For each TES state in the training set, we precompute action values with backward dynamic programming over a discretized storage state and convert them into deterministic rewards for RFT. This makes every candidate action verifiable under the same long-horizon objective used for control evaluation, so the model is trained against the relative quality of control decisions instead of against a single hard label or a human-preference signal. Figure~\ref{fig:planning_rft_framework} summarizes this training-and-inference framework.

From a learning perspective, the premise of this setting is that a common core of decision making runs through the entire task: every scheduling step poses the same kind of trade-off, only under different numbers. If RLVR can extract this core from the model's pretrained prior over multi-step decision making---in effect, a mild form of solver distillation \cite{hinton2015_distillation,anthony2017_expert_iteration}---then only a small number of verified examples should be needed. Reinforcement learning can be viewed as approximate dynamic programming: sampled estimates replace exact Bellman backups when the state--action space makes exhaustive DP infeasible \cite{bertsekas_tsitsiklis_ndp_1996,sutton_barto_rl_2018}. A sufficiently strong pretrained prior removes the need to rely exclusively on sampled approximations: when a few dozen verified states suffice, exact DP turns from an impractical teacher into a realistic one, and learning can be anchored directly to exact Bellman values rather than to their sampled surrogates. Backward DP is run once, offline, for the TES training states, and the resulting verifier dataset of state prompts and action values is amortized across all sampled completions and optimizer updates during RFT. Limiting training to a small set of verified states also makes the transfer test more informative: improvement on unseen states is harder to attribute to memorization of the training actions and provides stronger evidence that the model has learned a recurring decision pattern, supporting the generalization hypothesis in Section~\ref{subsec:gap-objectives}.

The division of labor also matters for deployment: the model-based component is used during training as a verifier, not during inference as an online controller. After post-training, the controller needs only structured state and forecast inputs, together with the action parser, and can in principle operate without an explicit online system model or repeated online optimization. To the extent that the learned decision procedure generalizes beyond a site-specific action table, this is the mechanism that would enable the cross-building transfer envisioned in Section~\ref{sec:introduction}.

\subsection{TES-specific training procedure}\label{subsec:rft-training-procedure}
\begin{figure}[t]
\centering
\includegraphics[width=\textwidth]{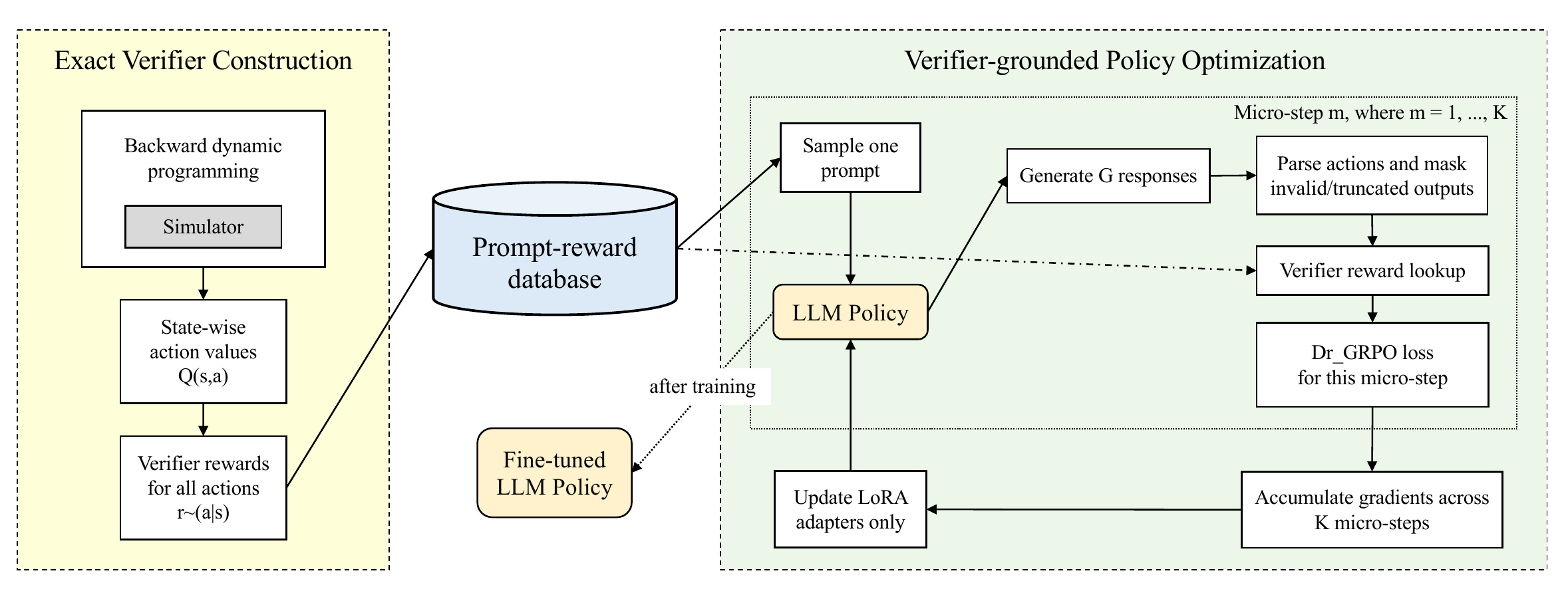}
\caption{TES-specific verifier construction and RFT training flow. The left panel shows offline verifier construction, and the right panel shows the online training loop. Here, $G$ denotes the number of generations per prompt and $K$ denotes the number of gradient-accumulation micro-steps per optimizer step.}
\label{fig:rft_training_flow}
\end{figure}

The TES-specific training pipeline, summarized in Figure~\ref{fig:rft_training_flow}, has two stages: offline verifier construction based on backward dynamic programming, and online policy optimization in which grouped model responses are scored by the verifier and used for RFT.

Offline, we construct the verifier by defining a finite-horizon dynamic program with the same objective as the TES control problem. The continuous TES energy is discretized into bins and the reduced DP state is represented as $(t,\bar E_t)$, where $\bar E_t$ is the discretized storage level and the external input trajectories are treated as known time-indexed inputs. Let $Q_t(\bar E_t,a)$ denote the action value for choosing action $a$ at state $(t,\bar E_t)$. With one-step reward $r_t(\bar E_t,a)$ and terminal reward $V_T(\bar E_T)$, the backward Bellman recursion is
\begin{equation}
Q_t(\bar E_t,a)=r_t(\bar E_t,a)+V_{t+1}(\bar E_{t+1}),
\end{equation}
\begin{equation}
V_t(\bar E_t)=\max_{a\in\mathcal{A}(t)} Q_t(\bar E_t,a).
\end{equation}
These equations use the standard finite-horizon Bellman recursion \cite{puterman_mdp_2005}. Here, $\mathcal{A}(t)$ is the admissible action set and $\bar E_{t+1}$ is obtained by applying the deterministic TES transition and mapping the resulting continuous storage level to the nearest bin. The backward recursion and terminal-state treatment follow standard dynamic-programming practice \cite{bertsekas_dpoc_vols_2012_2017}. To express the cost-minimization problem in reward form, the stage reward is defined as the negative of the stage cost and the terminal reward as the positive terminal value, so maximizing the DP return is equivalent to minimizing the control objective in Section~\ref{subsec:problem-formulation}. This yields a state-wise action value for every admissible decision under the same finite-horizon objective used in evaluation, which is exactly the quantity needed to make the RFT reward verifiable.

We write the precomputed value compactly as $Q(s_t,a)$ for action $a$ at state $s_t$. Rather than collapsing the verifier into a one-hot target that rewards only the single DP-optimal action, we retain the full vector of action values. This preserves graded differences in planning quality: when multiple actions have similar long-horizon value, near-optimal alternatives are not discarded as entirely wrong. Because the absolute scale of $Q$ varies across states and times, we convert it into state-normalized action rewards
\begin{equation}
\tilde r(a \mid s_t)=\frac{\exp(Q(s_t,a)/\tau)}{\sum_{a'\in\mathcal{A}(t)}\exp(Q(s_t,a')/\tau)},
\end{equation}
where $\tau$ is a temperature parameter. This yields a scalar verifier reward $\tilde r(a \mid s_t)\in(0,1)$ for each state-action pair; the rewards across actions sum to one, and the within-state ranking of actions is preserved. Smaller $\tau$ makes the signal closer to winner-take-all supervision, whereas larger $\tau$ retains more credit for near-optimal alternatives.

In the online training loop, the model is prompted with the same state representation used at inference time. For each prompt $x_t$ constructed from state $s_t$, $G$ responses $y^{(1)},\dots,y^{(G)}$ are sampled from the rollout policy $\pi_{\theta_{\mathrm{old}}}$, mapped to actions $a^{(i)}=g(y^{(i)})$, and assigned scalar rewards $R^{(i)}=\tilde r(a^{(i)} \mid s_t)$. Invalid outputs are masked from all subsequent computations so that optimization pressure is concentrated on valid control decisions; let $\mathcal{V} \subseteq \{1,\dots,G\}$ denote the index set of valid responses.
\begin{algorithm}[!t]
\small
\SetAlgoLined
\DontPrintSemicolon
\caption{Verifier-based RFT for TES control}
\label{alg:tes_rft}
\KwIn{TES training states and prompts $\mathcal{S}_{\mathrm{train}}=\{(s,x_s)\}$, TES DP model, temperature $\tau$, policy $\pi_\theta$, parser $g(\cdot)$, group size $G$, accumulation length $K$, maximum number of optimizer steps $N_{\mathrm{opt}}$}
\KwOut{Fine-tuned TES policy $\pi_\theta$}

Run backward DP under the TES dynamics and objective to compute $Q(s,a)$ for each training state and construct $\mathcal{D}=\{(x_s,Q(s,\cdot))\}$\;
Offline, convert each $Q(s,\cdot)$ in $\mathcal{D}$ into state-normalized action rewards $\tilde r(\cdot \mid s)$ by Boltzmann normalization\;
Build a prompt stream that repeatedly samples one TES prompt and replicates it $G$ times\;
\For{$n \leftarrow 1$ \KwTo $N_{\mathrm{opt}}$}{
Clear accumulated gradients\;
\For{$m \leftarrow 1$ \KwTo $K$}{
Sample one prompt group $(x_s,\tilde r(\cdot \mid s))$ from the stream\;
Generate $G$ responses $y^{(1)},\ldots,y^{(G)}$ for the same prompt with the current rollout policy\;
\For{$i \leftarrow 1$ \KwTo $G$}{
Parse action $a^{(i)} \leftarrow g(y^{(i)})$\;
\eIf{$a^{(i)}$ is invalid or truncated}{
Mark reward as NaN so the sample is masked from optimization\;
}{
Assign scalar reward $R^{(i)} \leftarrow \tilde r(a^{(i)} \mid s)$\;
}
}
Compute centered group-relative advantages $\hat A^{(i)} = R^{(i)} - \mu_R$ over the valid responses\;
Accumulate gradients for one micro-step with the Dr\_GRPO objective over the valid responses from the same prompt\;
}
Apply one optimizer step to update only the LoRA parameters of $\pi_\theta$\;
}
\end{algorithm}

Policy updates use the group-relative policy-optimization formulation introduced with DeepSeekMath \cite{deepseekmath2024}. This objective compares multiple generations from the same prompt and shifts probability mass toward higher-reward actions; subsequent analysis characterizes the effective loss and success-amplification dynamics of this update \cite{mroueh2025rlvr}. Its clipped importance-ratio surrogate is inherited from PPO \cite{ppo2017}. Let
\begin{equation}
\mu_R=\frac{1}{|\mathcal{V}|}\sum_{i\in\mathcal{V}} R^{(i)}
\end{equation}
be the mean reward over the valid responses in the group. Following the centered group-relative formulation adopted here \cite{liu2025r1zero}, the advantage of each valid sample $i\in\mathcal{V}$ is
\begin{equation}
\hat A^{(i)}=R^{(i)}-\mu_R.
\end{equation}
For token $u$ in sampled response $y^{(i)}$, the importance ratio is
\begin{equation}
\rho^{(i)}_{u}(\theta)=
\frac{\pi_{\theta}(y^{(i)}_{u}\mid x_t,y^{(i)}_{<u})}
{\pi_{\theta_{\mathrm{old}}}(y^{(i)}_{u}\mid x_t,y^{(i)}_{<u})},
\end{equation}
and the clipped surrogate term is
\begin{equation}
\ell^{(i)}_{u}(\theta)=
\min\!\left(
\rho^{(i)}_{u}(\theta)\hat A^{(i)},
\mathrm{clip}\!\left(\rho^{(i)}_{u}(\theta),1-\epsilon,1+\epsilon\right)\hat A^{(i)}
\right),
\end{equation}
where $\epsilon$ is the clipping parameter. We use the Dr\_GRPO variant, which replaces response-length normalization with a fixed maximum completion length $L_{\max}$ \cite{liu2025r1zero}:
\begin{equation}
\mathcal{J}_{\mathrm{Dr\_GRPO}}(\theta)=
\mathbb{E}\!\left[
\frac{1}{|\mathcal{V}|}\sum_{i\in\mathcal{V}}\frac{1}{L_{\max}}\sum_{u=1}^{|y^{(i)}|}\ell^{(i)}_{u}(\theta)
\right].
\end{equation}
Maximizing $\mathcal{J}_{\mathrm{Dr\_GRPO}}$ increases the probability of token sequences associated with better TES actions relative to other candidates from the same state. Parameter-efficient adaptation is introduced through Low-Rank Adaptation (LoRA) \cite{lora2022}, so that the base model remains fixed and only lightweight adapters are updated.

The prompt structure is also kept consistent between training and evaluation: the fine-tuned model sees the same decomposition into role definition, output constraint, and state description that it later encounters at test time. This allows gains after RFT to be interpreted as changes in the model's decision procedure under a stable control interface, not as artifacts of a different prompt template. It also motivates a broader evaluation question beyond nominal TES performance, following the hypothesis in Section~\ref{subsec:gap-objectives}: whether the decision patterns encouraged by the verifier remain effective under shifted TES conditions and in a related storage-control task without additional task-specific adaptation.

Algorithm~\ref{alg:tes_rft} summarizes the complete verifier-based RFT procedure used in the TES experiments.

\section{Experimental Setup}\label{sec:experimental-setup}
\subsection{TES control problem formulation}\label{subsec:problem-formulation}
\begin{figure}[t]
\centering
\includegraphics[width=\textwidth]{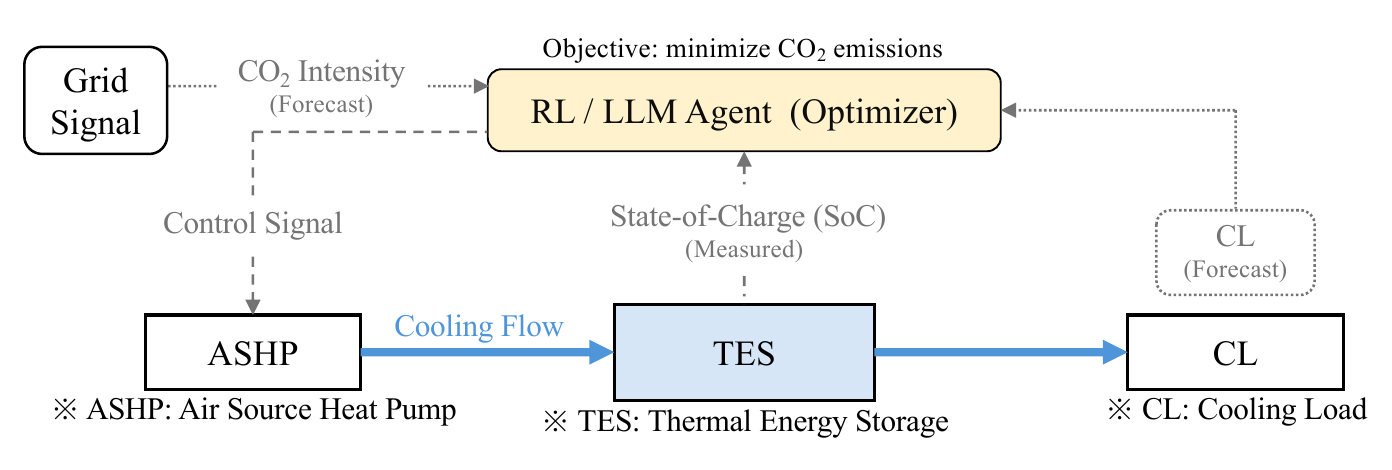}
\caption{System diagram of the TES control problem. The TES tank is arranged as flow-through storage between the ASHP and the cooling load: all cooling passes through the tank, which absorbs the hourly imbalance between ASHP output and demand, and same-hour direct supply in the formulation corresponds to flow-through of the tank. The RL/LLM agent receives the current TES state, cooling-load forecast, and grid carbon-intensity forecast, and then selects the ASHP control action to minimize cumulative CO$_2$ emissions.}
\label{fig:tes_system_diagram}
\end{figure}

We formulate TES scheduling as a finite-horizon discrete-time Markov decision process (MDP) \cite{puterman_mdp_2005}. At each hour $t$, the controller selects the output level of an air-source heat pump (ASHP) that supplies cooling demand directly and, when possible, charges a water-based TES tank for later use. The decision depends on cooling demand, grid carbon intensity, and the current storage level. Figure~\ref{fig:tes_system_diagram} summarizes this control loop and the information flow between the optimizer, the TES system, and the external inputs. Our TES control setting is motivated by prior carbon-aware TES operation studies, including Miyata et al.~\cite{miyata2020_dynamic_co2_mpc}. We use a simplified but physically consistent benchmark model to evaluate the proposed control method under explicit storage and demand constraints.

The action at time $t$ is the ASHP output ratio
\begin{equation}
a_t \in \left\{0,\frac{1}{3},\frac{2}{3},1\right\},
\end{equation}
which determines the cooling output
\begin{equation}
Q_t^{\mathrm{ch}} = a_t Q^{\max}_{\mathrm{ch}},
\end{equation}
where $Q^{\max}_{\mathrm{ch}}$ is the rated cooling capacity. Given cooling demand $L_t$ and TES energy $E_t$, direct supply from the ASHP is
\begin{equation}
Q_t^{\mathrm{dir}}=\min(L_t,Q_t^{\mathrm{ch}}),
\end{equation}
the remaining demand is $R_t=L_t-Q_t^{\mathrm{dir}}$, and TES discharge is
\begin{equation}
Q_t^{\mathrm{dis}}=\min(R_t,E_t).
\end{equation}
Any unmet demand is represented explicitly as
\begin{equation}
U_t=\max(0,R_t-Q_t^{\mathrm{dis}}),
\end{equation}
which allows all methods to be compared under a shared soft-constraint formulation. If ASHP output exceeds immediate demand, the surplus
\begin{equation}
S_t=\max(0,Q_t^{\mathrm{ch}}-Q_t^{\mathrm{dir}})
\end{equation}
is used to charge TES up to the available capacity,
\begin{equation}
Q_t^{\mathrm{cha}}=\min(S_t,E^{\max}-E_t),
\end{equation}
and the storage state is updated as
\begin{equation}
E_{t+1}=\mathrm{clip}\left(E_t-Q_t^{\mathrm{dis}}+Q_t^{\mathrm{cha}},0,E^{\max}\right).
\end{equation}

Because the control interval is $\Delta t=1$~h, power in kilowatts and per-interval energy in kilowatt-hours are numerically identical, and $\Delta t$ is omitted for brevity. These relations describe a tank-buffered cooling system in which the chiller operates against a water storage tank that buffers energy between the heat source and the load, as in the system studied by Miyata et al.~\cite{miyata2020_dynamic_co2_mpc}. As shown in Fig.~\ref{fig:tes_system_diagram}, the TES tank is accordingly arranged as flow-through storage between the ASHP and the cooling load, hydraulically decoupling the heat-source side from the load side. The decomposition into direct supply, discharge, and charging above is an accounting convention for this single tank balance: the combined update reduces to $E_{t+1}=\mathrm{clip}\left(E_t+Q_t^{\mathrm{ch}}-L_t,\,0,\,E^{\max}\right)$, so the tank absorbs the hourly imbalance between ASHP output and demand, and same-hour ``direct supply'' corresponds to water passing through the tank. Because the benchmark tank is lossless, the two descriptions are exactly equivalent.

The ASHP electricity consumption, using the coefficient of performance (COP), is
\begin{equation}
P_t=\frac{Q_t^{\mathrm{ch}}}{\mathrm{COP}},
\end{equation}
and stepwise carbon emissions are
\begin{equation}
C_t=P_t g_t,
\end{equation}
where $g_t$ is the dynamic grid carbon-intensity signal.

The controller minimizes cumulative operating emissions while strongly penalizing unmet cooling demand. Accordingly, the trajectory cost is defined as
\begin{equation}
J=\sum_{t=0}^{T-1}\left(C_t+\beta U_t\right)-V_{\mathrm{term}}(E_T),
\end{equation}
where $\beta$ is a large penalty weight and $V_{\mathrm{term}}(E_T)$ is a terminal value that approximates the benefit of carrying useful TES energy into the next day \cite{bertsekas_dpoc_vols_2012_2017}. In the experiments, this terminal value is implemented as
\begin{equation}
V_{\mathrm{term}}(E_T)=\bar g^{\mathrm{low}}_{\mathrm{next}}\frac{E_T}{\mathrm{COP}},
\end{equation}
where $\bar g^{\mathrm{low}}_{\mathrm{next}}$ is the average of the five lowest next-day operating-hour carbon-intensity values. Under the nominal perfect-forecast assumption, the problem is Markovian when the state includes the current storage level, time index, and the forecast information needed for planning. For the DP verifier, the demand and carbon-intensity trajectories are treated as known time-indexed inputs, so the reduced state $(t,E_t)$ is sufficient. For RL and LLM controllers, we instead provide the remaining same-day forecasts and next-day summary statistics in the observation. This shared formulation allows performance differences to be attributed to decision quality rather than to mismatched physics or evaluation rules.

\subsection{TES simulation environment and operating conditions}\label{subsec:tes-setup}
The nominal TES test case represents a water-based cooling system with thermal storage for a Tokyo office building; all methods are evaluated in the same simulator. Time is discretized at 1-h resolution, and each evaluation episode spans three consecutive days ($T=72$ steps). Because the terminal value depends on next-day carbon conditions, one additional day of external input data is used only to compute next-day summary statistics and the end-of-horizon storage value. In the nominal experiments, the main comparison window is May 5--7, 2025.

The physical setting follows the TES formulation in Section~\ref{subsec:problem-formulation}. The ASHP has a rated cooling capacity of 100~kW and a constant COP of 4.0, and the TES capacity is 300~kWh. Each evaluation episode starts with the storage at 50\% of capacity ($E_0=150$~kWh). Cooling demand is assumed to occur only during business hours, so active control decisions are made for hourly intervals starting at 08:00,\dots,17:00, while the ASHP is kept off outside that window. Dynamic carbon intensity is computed from area-level power-system data and aggregated into the hourly signal used by the simulator. Under nominal conditions, the controller receives perfect same-day forecasts of cooling demand and carbon intensity, together with compact next-day summary statistics for the terminal-value approximation. The unmet-demand penalty weight is fixed at $\beta=10$ throughout the TES experiments. Table~\ref{tab:tes_environment} summarizes these settings. Consistent with the scope set in the Introduction, the environment omits COP variation and storage heat-loss dynamics by design (Section~\ref{subsec:limitations}).

To keep the comparison between numerical controllers and LLMs fair, RL and LLM controllers receive the same decision information in different formats. The shared observation contains the current TES level, time information, current demand and carbon intensity, the remaining same-day forecast trajectory for operating hours, and three next-day summary features: mean operating-hour demand, minimum operating-hour carbon intensity, and the average of the five lowest next-day operating-hour carbon-intensity values. Under nominal conditions, this provides a Markovian control interface while remaining compact enough for both numerical RL inputs and text prompts. The open-weight model used for RFT is trained on prompts constructed from three days of TES trajectories in April 2025, corresponding to 30 training prompts (one per active decision hour), so that the main May 5--7, 2025 test episode remains out of sample.

\begin{table}[t]
\centering
\caption{Nominal TES environment and evaluation settings.}
\label{tab:tes_environment}
\small
\begin{tabular}{ll}
\toprule
Item & Setting \\
\midrule
Time step & 1 h \\
Evaluation horizon & 3 d ($T=72$) + 1 d for terminal statistics \\
Control window & 08:00--18:00 business hours \\
Action set & $\{0,\frac{1}{3},\frac{2}{3},1\}$ ASHP output ratio \\
ASHP rated cooling capacity & 100 kW \\
ASHP COP & 4.0 \\
TES capacity & 300 kWh \\
Initial TES energy $E_0$ & 150 kWh (50\% of capacity) \\
Primary test period & May 5--7, 2025 \\
Primary objective & $\sum_t (C_t+\beta U_t)-V_{\mathrm{term}}(E_T)$, $\beta=10$ \\
\bottomrule
\end{tabular}
\end{table}
\subsection{TES-specific RFT training configuration}\label{subsec:rft-training-config}
For the TES experiments, we use a compact, fixed training configuration instead of a task-by-task hyperparameter search, and the main hyperparameters are summarized in Table~\ref{tab:tes_rft_hparams}. The most consequential settings are the group size $G=16$, LoRA-only adaptation of the attention projection layers, the Dr\_GRPO objective, and the reward temperature $\tau=0.05$ used to convert DP action values into state-normalized action rewards. TES-specific RFT was conducted on a single NVIDIA H100 SXM 80GB GPU.

\begin{table}[t]
\centering
\caption{Main hyperparameters for TES-specific RFT.}
\label{tab:tes_rft_hparams}
\small
\begin{tabular}{ll}
\toprule
Item & Setting \\
\midrule
Base model & \texttt{openai/gpt-oss-20b} \\
LoRA target modules & \begin{tabular}[t]{@{}l@{}}\texttt{q\_proj}, \texttt{k\_proj},\\ \texttt{v\_proj}, \texttt{o\_proj}\end{tabular} \\
LoRA rank $r$ / scaling $\alpha$ & $4$ / $8$ \\
Learning rate & $5\times10^{-5}$ \\
Training steps & $200$ \\
Group size $G$ (samples per prompt) & $16$ \\
Prompts per micro-step & $1$ \\
Per-device training batch size & $16$ \\
Gradient accumulation steps & $4$ \\
Maximum prompt length & $1000$ \\
Maximum completion length $L_{\max}$ & $4000$ \\
Generation temperature / top-$p$ & $1.0$ / $1.0$ \\
Loss & Dr\_GRPO \\
Reward temperature $\tau$ & $0.05$ \\
\bottomrule
\end{tabular}
\end{table}
\subsection{Baselines and comparison settings}\label{subsec:baselines}
We compare the controller settings summarized in Table~\ref{tab:tes_controller_settings} under the shared TES dynamics. A no-TES baseline is included to quantify the benefit of thermal storage itself: in this setting, the ASHP directly follows the cooling demand without storing surplus cooling energy. DP serves as the optimal reference because it solves the finite-horizon problem with backward recursion \cite{puterman_mdp_2005}. MPC is implemented as a receding-horizon controller on the same benchmark dynamics used by all other methods, with a maximum 36-h look-ahead window and differential-evolution-based approximate action-sequence optimization at each decision step \cite{ref:storn_price_de_1997}. The model-free RL comparison includes Proximal Policy Optimization (PPO) \cite{ppo2017} and a Deep Q-Network (DQN) \cite{ref:mnih2015_dqn_nature}. Both are trained in the same simulator using April--June 2024 TES episodes and the same state information as the LLM controllers, represented as numerical vectors instead of text. In the main comparison, DQN is reported because it achieved lower nominal emissions than PPO under the same protocol. Appendix~\ref{app:baseline-hparams} lists the main DP, MPC, and RL settings.

The LLM entries have distinct roles. GPT-4o is included as a strong general-purpose non-reasoning LLM baseline, so that its comparison with GPT-5 can help assess whether deliberate multi-step planning contributes to TES control quality beyond instruction following alone. GPT-5 serves as the frontier reasoning-model reference \cite{ref:gpt5_system_card}. The open-weight model \texttt{gpt-oss-20b} is evaluated both before and after TES-specific RFT to isolate the effect of post-training on decision quality and reasoning traces. The closed-model baselines were evaluated via the OpenAI API in August 2025, using \texttt{gpt-5-2025-08-07} with medium reasoning effort and \texttt{gpt-4o-2024-08-06}. Sampling temperature was not explicitly specified and was left at the API default.

\begin{table}[t]
\centering
\caption{Controller settings used in the nominal TES comparison.}
\label{tab:tes_controller_settings}
\small
\begin{tabular}{p{0.18\textwidth}p{0.28\textwidth}p{0.43\textwidth}}
\toprule
Controller setting & Role in comparison & Implementation and input interface \\
\midrule
No-TES baseline & Storage-free reference & ASHP directly follows cooling demand; surplus cooling is not stored. \\
DP & Optimal reference & Backward dynamic programming over a discretized TES-energy state with known future trajectories. \\
MPC & Conventional optimization baseline & Receding-horizon optimization on the shared benchmark dynamics with a maximum 36-h look-ahead window. \\
RL & Model-free learning baseline & PPO and DQN are trained with the same observation information as the LLM controllers in numerical vector form; DQN is reported because it achieved lower nominal emissions than PPO under the same protocol. \\
GPT-4o & Non-reasoning general-purpose LLM baseline & Prompt-based controller using the shared textual TES observation and strict action parsing. \\
GPT-5 & Frontier reasoning-model reference & Prompt-based controller using the shared textual TES observation and a JSON action output with a brief logged reasoning summary. \\
\texttt{gpt-oss-20b} Pre-RFT & Open-weight model before task adaptation & Prompt-based controller evaluated before TES-specific reinforcement fine-tuning. \\
\texttt{gpt-oss-20b} Post-RFT & Open-weight model after task adaptation & Same open-weight controller after DP-verifier-based RFT using TES training prompts. \\
\bottomrule
\end{tabular}
\end{table}

All LLM controllers use the prompt decomposition and output constraints described in Section~\ref{subsec:controller-design}, and only the parsed action is executed. No tools, retrieval modules, or external memory are used. For the optimization and learning baselines, the same system model and objective are used, but information access differs in the standard way: DP and MPC optimize directly over future numerical trajectories, whereas RL and LLM controllers receive the fixed observation design described above.

The primary evaluation metric is the shared finite-horizon objective $J$ in Section~\ref{subsec:problem-formulation}. Unmet cooling demand is tracked explicitly and was zero in every run reported in the result tables. The penalty term therefore vanishes, and the objective reduces to cumulative operating emissions minus the terminal value $V_{\mathrm{term}}(E_T)$. We report this terminal-value-adjusted emissions objective in kg-CO$_2$ and call it total emissions for brevity. The only case with unmet demand (MPC under the forecast-error stress test) is not included in these tables and is discussed in Section~\ref{subsec:robustness}. For the open-weight model, we additionally analyze model-generated reasoning outputs as a secondary behavioral comparison before and after RFT. The setting and results of the forecast-error stress test are presented together in Section~\ref{subsec:robustness}.

\section{TES Control Results}\label{sec:tes-results}
\subsection{Overall control performance}\label{subsec:overall-performance}
Table~\ref{tab:tes_nominal_results} summarizes the nominal TES results under the shared evaluation protocol. The central result is the effect of verifier-based RFT: TES-specific post-training reduces the emissions of the open-weight model from 70.5 to 61.2 kg-CO$_2$, moving it from behind the RL baseline to within 0.4 kg-CO$_2$ of the DP reference---a gain achieved after training on only 30 prompts from three days in April 2025 and observed on a temporally held-out test episode in May 2025 (Section~\ref{subsec:tes-setup}). The other controllers place this result in context. DP, MPC, and GPT-5 form the reference tier, with GPT-5 reaching essentially the same emission level as the optimization references without task-specific training. The representative RL baseline (DQN; Section~\ref{subsec:baselines}) improves substantially over the no-TES baseline but remains behind this tier. GPT-4o performs worst and even underperforms the no-TES baseline. Access to storage alone is not enough: without strong multi-step planning, a controller can use TES in ways that increase rather than reduce emissions.

\begin{figure}[t]
\centering
\begin{minipage}[t]{0.47\textwidth}
\centering
\includegraphics[width=\textwidth]{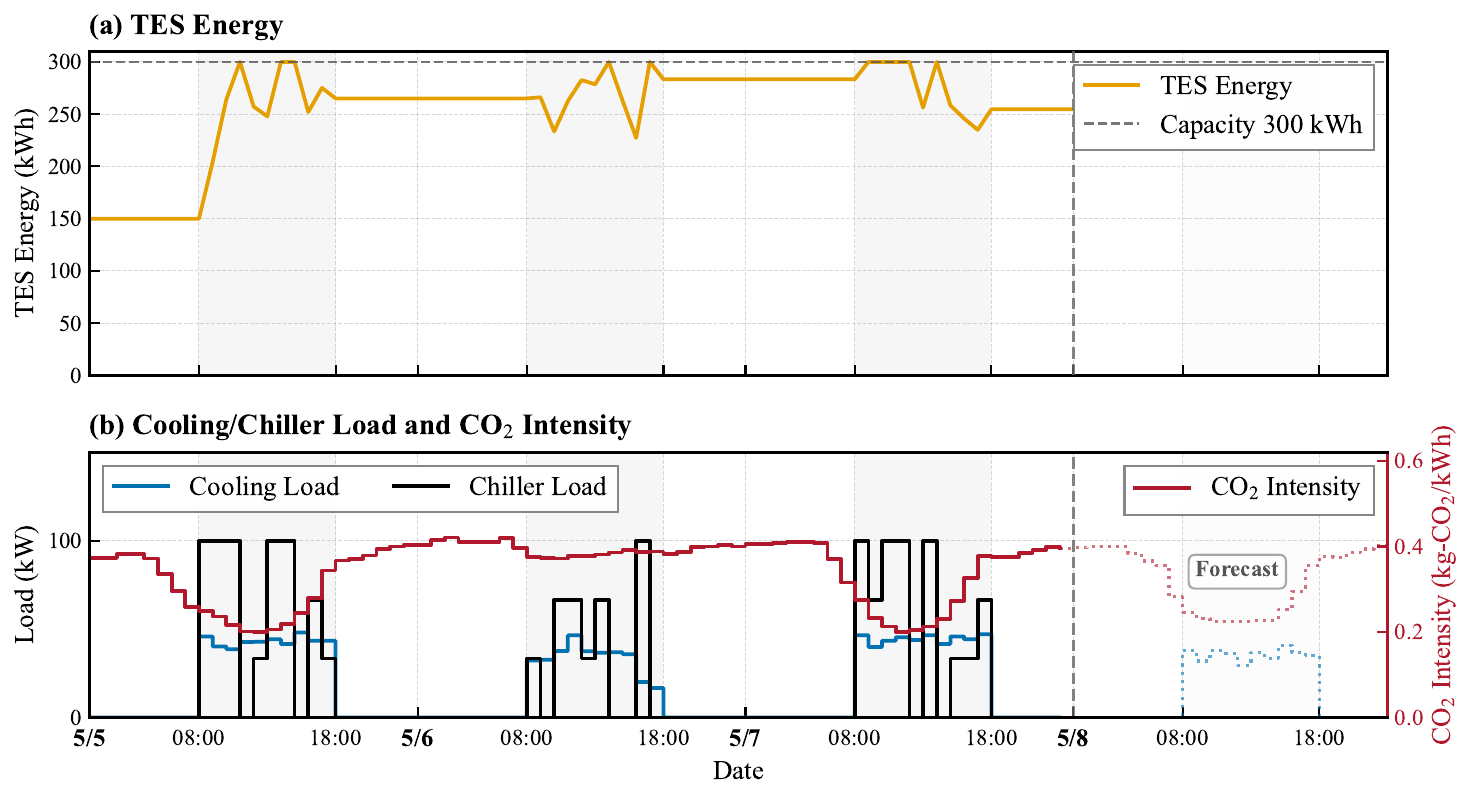}

{\small (a) GPT-4o}
\end{minipage}\hfill
\begin{minipage}[t]{0.47\textwidth}
\centering
\includegraphics[width=\textwidth]{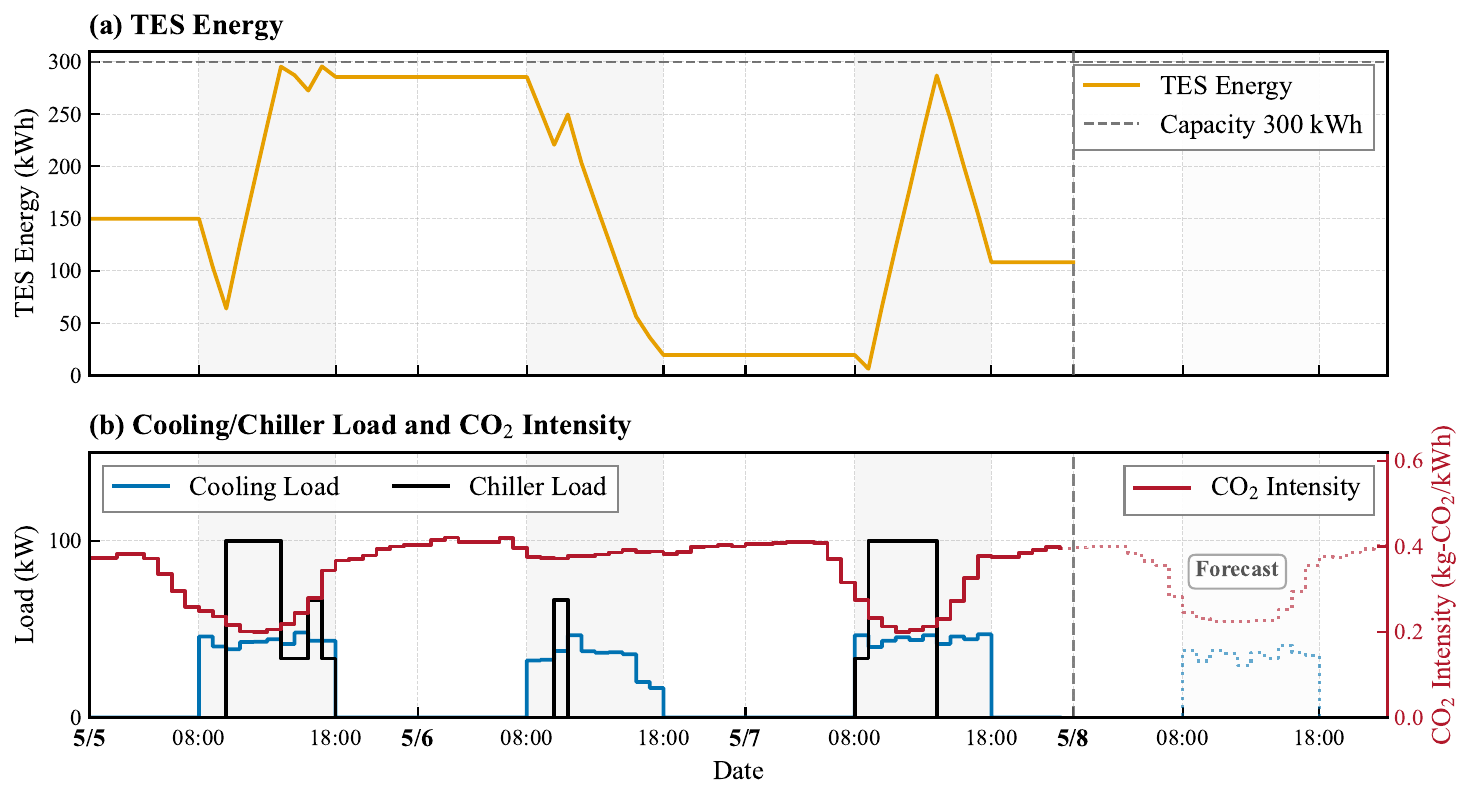}

{\small (b) GPT-5}
\end{minipage}

\vspace{0.8em}

\begin{minipage}[t]{0.47\textwidth}
\centering
\includegraphics[width=\textwidth]{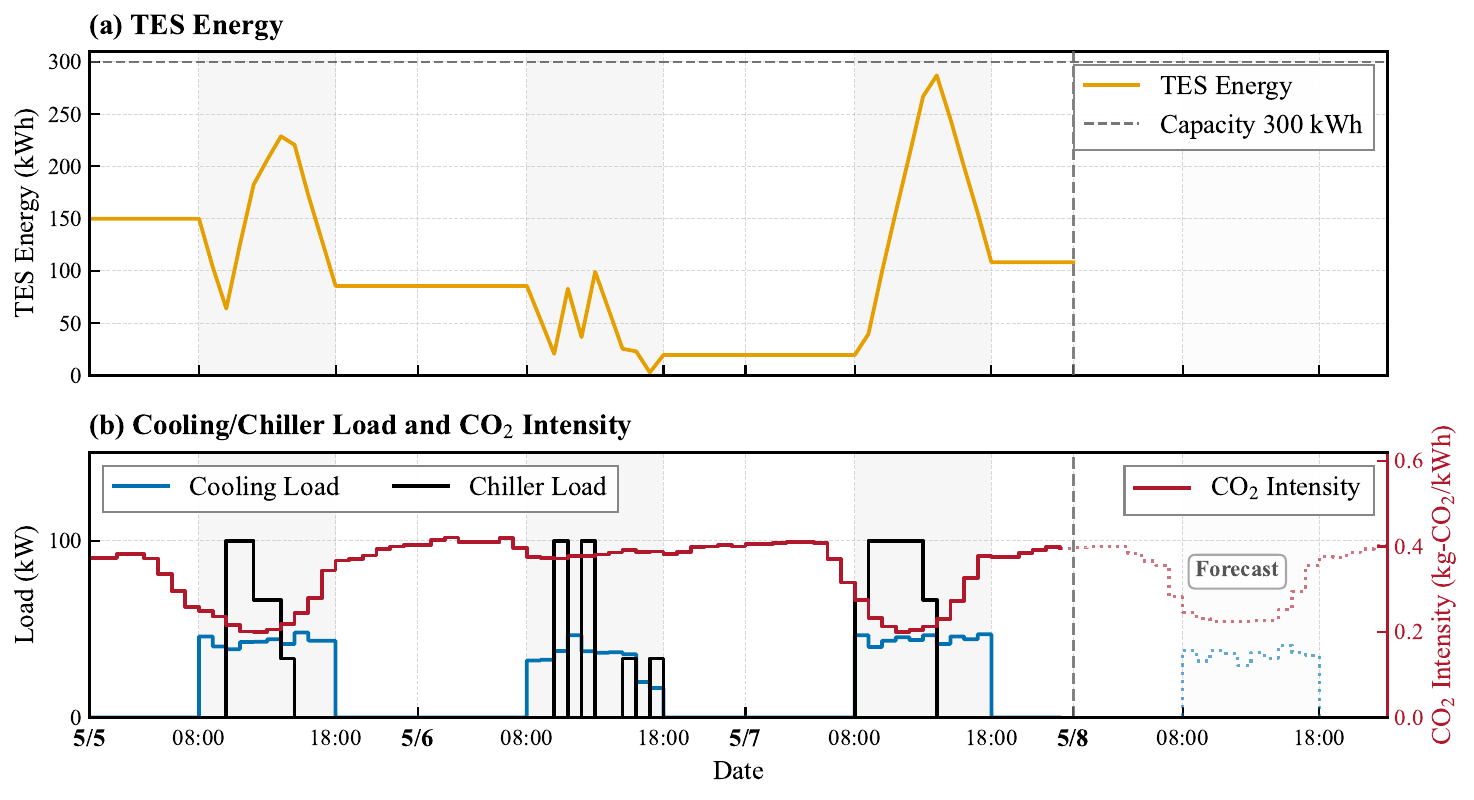}

{\small (c) RL (DQN)}
\end{minipage}\hfill
\begin{minipage}[t]{0.47\textwidth}
\centering
\includegraphics[width=\textwidth]{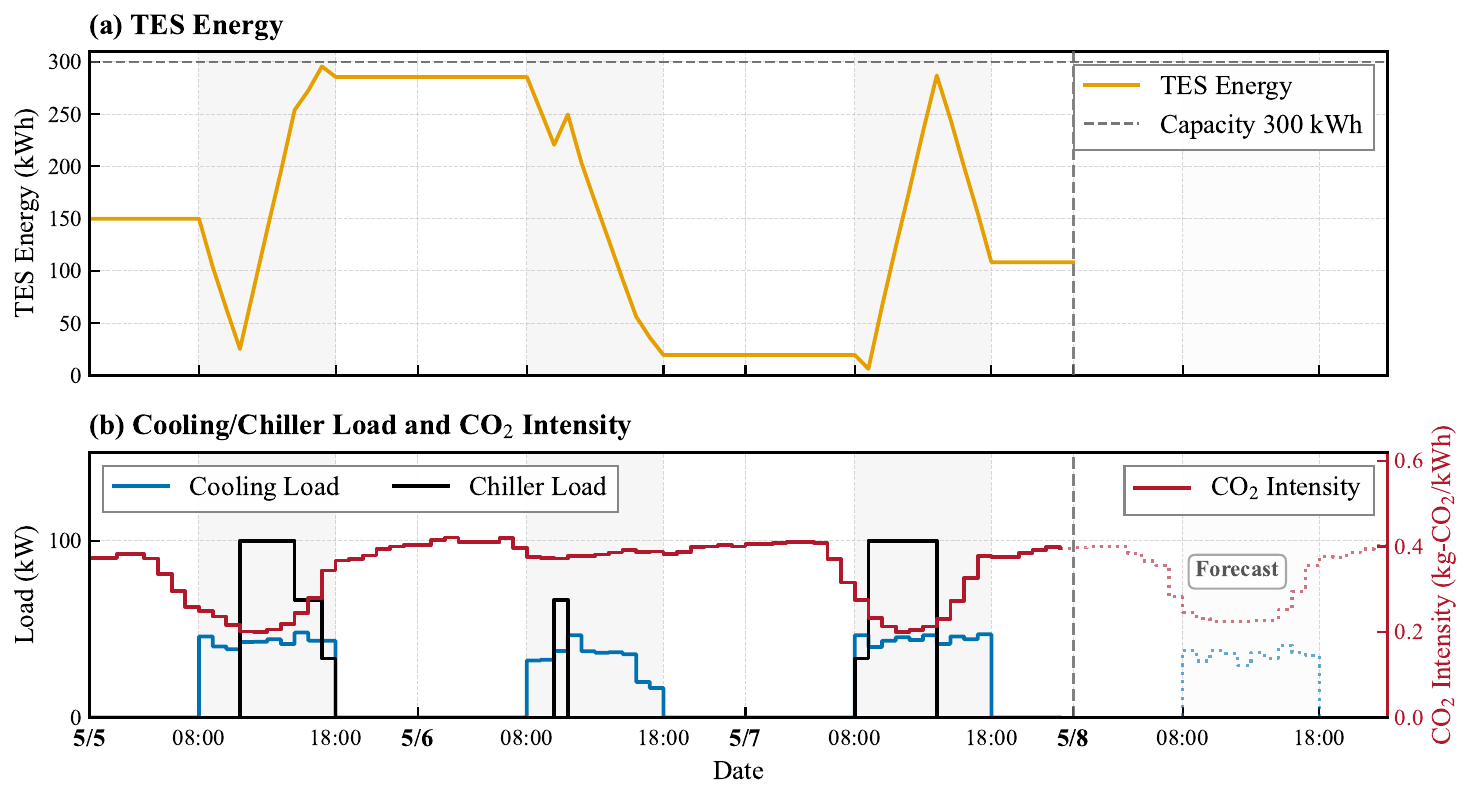}

{\small (d) gpt-oss-20b Post-RFT}
\end{minipage}
\caption{Representative nominal TES trajectories for selected controllers. In each panel, the upper subplot presents the TES energy trajectory [kWh], while the lower subplot presents cooling load and chiller load [kW] together with grid CO$_2$ intensity [kg-CO$_2$/kWh].}
\label{fig:tes_nominal_llm}
\end{figure}

The poorer performance of GPT-4o also suggests that the benchmark, although built to be tractable, is not trivial: it distinguishes multi-step planning from fluent instruction following. The representative RL baseline does learn the basic low-carbon charging/high-carbon discharging pattern, as shown in Fig.~\ref{fig:tes_nominal_llm}, but it does not reliably recover the same longer-horizon coordination among current TES headroom, intraday carbon minima, and end-of-day storage value. The RL gap says more about the parameter budget and training data used here than about any intrinsic unsuitability of model-free RL. This reading is consistent with broader deep-RL evidence: control performance depends heavily on learned representations \cite{ref:mnih2015_dqn_nature}, and generalization on the diversity of training environments \cite{ref:cobbe2019_rl_generalization}.

The nominal trajectories in Fig.~\ref{fig:tes_nominal_llm} further illustrate this pattern. The post-RFT open-weight model avoids the large early charging errors observed before RFT and approaches the qualitative behavior of the optimization references: it preserves TES headroom for cleaner charging periods and then discharges during dirtier periods while staying within the storage bounds. GPT-5 illustrates near-optimal performance on the same episode: it closely reproduces the DP-like schedule across the full three-day window, and its action sequence differs from DP at only one active step, on day~3 at 09:00, with a negligible cumulative-emission difference. A representative GPT-5 reasoning summary expresses the underlying planning logic concisely:
\begin{quote}
\small\itshape
``CO$_2$ is lower later ... Turning the ASHP off now discharges TES, creates headroom to max-charge at midday, and preserves stored energy for high-CO$_2$ hours.''
\end{quote}
Even in this short summary-style format, the model explicitly links the current action to three forward-looking considerations that are central to TES scheduling: use TES now, keep capacity available for cleaner intraday charging, and retain stored cooling for dirtier later periods.

\begin{table}[t]
\centering
\caption{Nominal TES control results under the shared evaluation setting.}
\label{tab:tes_nominal_results}
\small
\begin{tabular}{l@{\hspace{2.5em}}r}
\toprule
Controller & Total emissions [kg-CO$_2$] \\
\midrule
GPT-4o & 95.4 \\
Baseline (No Thermal Storage) & 85.9 \\
\mbox{gpt-oss-20b} Pre-RFT & 70.5 \\
RL (DQN) & 67.5 \\
\mbox{gpt-oss-20b} Post-RFT & 61.2 \\
GPT-5 & 60.9 \\
MPC & 60.9 \\
DP & 60.8 \\
\bottomrule
\end{tabular}
\end{table}

\subsection{Analysis of model-generated reasoning traces}\label{subsec:trace-analysis}
Across many nominal TES decisions, both versions of \texttt{gpt-oss-20b} follow the same broad five-stage sequence: problem restatement (R1), inspection of the current and forecast context (R2), generation of candidate actions (R3), forward simulation of consequences (R4), and final constraint checking (R5). We code the traces only at the level of these externally visible operations and do not assume that they reveal the model's internal computation \cite{chen2025reasoning_models}. A reasoning process is considered stronger post-RFT when it more reliably exhibits R3--R5 in states where good control requires explicit comparison, look-ahead, and feasibility checking. The evidence below identifies where this sequence becomes more reliable after RFT.

\begin{figure}[t]
\centering
\begin{minipage}[t]{0.48\textwidth}
\centering
\includegraphics[width=\textwidth,height=0.24\textheight,keepaspectratio]{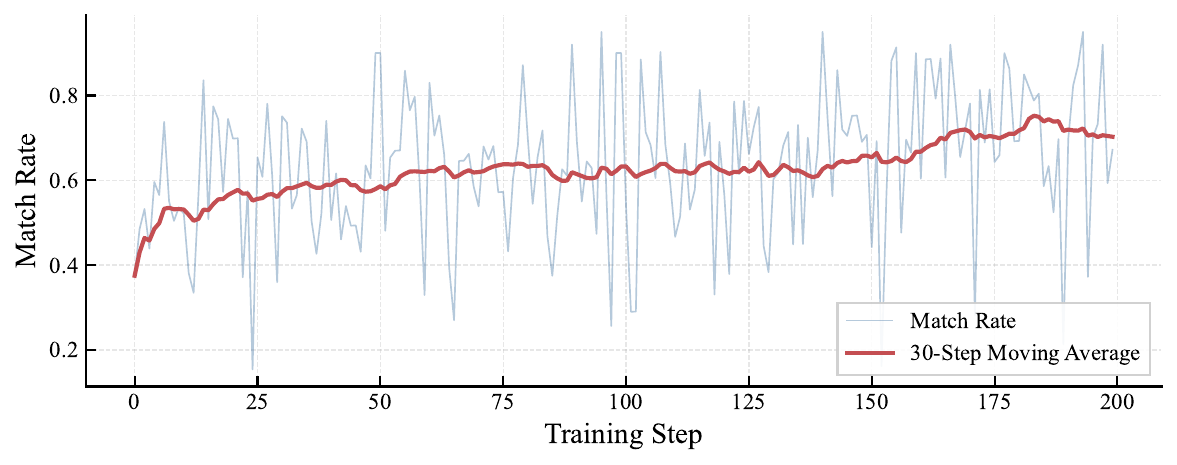}

{\small (a) DP action-match rate during RFT}
\end{minipage}\hfill
\begin{minipage}[t]{0.48\textwidth}
\centering
\includegraphics[width=\textwidth,height=0.24\textheight,keepaspectratio]{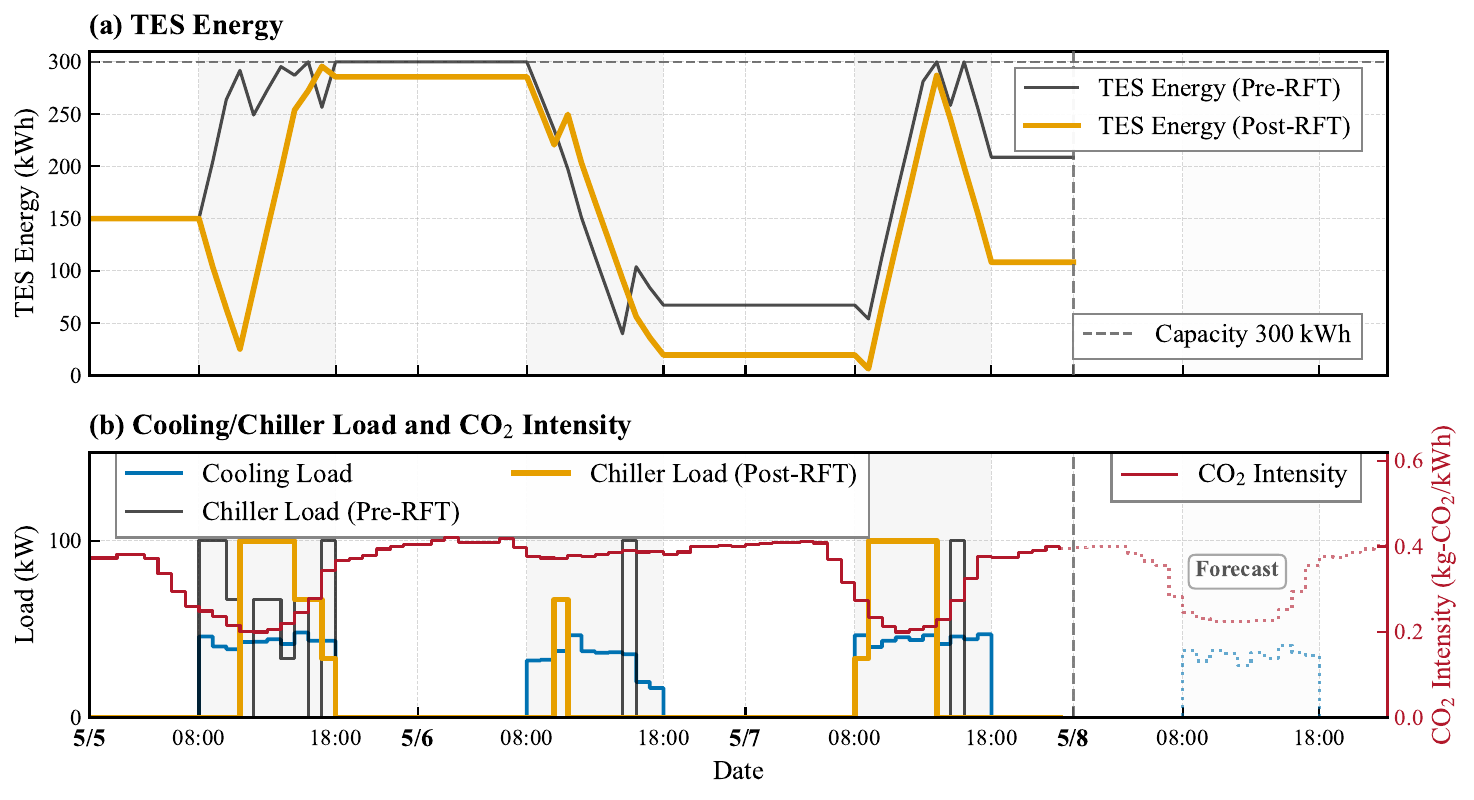}

{\small (b) Before/after comparison of \texttt{gpt-oss-20b}}
\end{minipage}
\caption{TES reasoning-process analysis results. Panel (a) shows the DP action-match rate during RFT, and panel (b) shows the TES control trajectories before and after RFT.}
\label{fig:rft_trace_shift}
\end{figure}

Figure~\ref{fig:rft_trace_shift} summarizes two complementary indicators of the effect of TES-specific RFT. First, the action-match rate against the DP reference rises steadily during training, indicating that the model increasingly reproduces the action pattern favored by the verifier. Second, the before/after control comparison shows that the post-RFT model corrects a characteristic pre-RFT error at the beginning of the horizon. Before RFT, the model often commits too early to charging because it gives too much weight to coarse next-day summary statistics relative to within-day carbon minima. After RFT, the model more often compares explicit candidates and re-checks the same decision, preserving headroom and delaying charging until cleaner midday periods when that is globally preferable.

The same shift is visible in short trace excerpts. In a representative Pre-RFT trace, the model places too much weight on the next-day forecast context and selects a charging action without adequately reconsidering the cleaner within-day trend:
\begin{quote}
\small\itshape
``Here at 08:00, CO$_2$ 0.249, lower than tomorrow ... So we should store as much as possible today to use tomorrow.''
\end{quote}
By contrast, a representative Post-RFT trace externalizes a more deliberate compare-and-verify loop:
\begin{quote}
\small\itshape
``Let's consider two main options \ldots''\\
``But let's verify this against future opportunities \ldots''
\end{quote}
After RFT, the model-generated traces more frequently exhibit explicit candidate comparison, forward simulation, and feasibility checking, and more often enumerate multiple candidate actions before committing to one. As an additional check, we performed a best-of-10 analysis over the same three-day evaluation period used in Section~\ref{subsec:overall-performance}: when ten candidate responses were sampled at each decision step and the candidate matching the DP reference action was selected whenever available, the Pre-RFT and Post-RFT open-weight models produced the same TES control trajectory. The near-optimal action sequence was therefore already reachable within the Pre-RFT sampling distribution; TES-specific RFT primarily increases the probability of selecting such actions and does not introduce a new control strategy. In energy control, where decisions must be reliably near-optimal rather than occasionally optimal, this improvement in consistency is operationally meaningful. Together with the performance gain on the held-out test episode, this diagnostic is consistent with the working hypothesis in Section~\ref{subsec:gap-objectives}: RFT stabilizes a reusable scheduling procedure instead of memorizing case-specific actions. The next section tests the generalization component of this hypothesis.

\section{Robustness and Generalization Tests}\label{sec:generalization-tests}
The DP verifier used for RFT scored candidate actions only under the nominal TES condition, with demand and carbon-intensity trajectories treated as known inputs. This section tests the limits of the reinforced behavior under progressively larger departures from that condition: introducing errors into the forecast information available to the controller (Section~\ref{subsec:robustness}), shifting the numerical regime through changes in storage capacity and regional conditions (Section~\ref{subsec:capacity-region}), and changing the physical semantics through a battery-scheduling task (Section~\ref{subsec:battery-transfer}). Section~\ref{subsec:transfer-limits} then identifies which behaviors persist.

\subsection{Forecast-error stress test}\label{subsec:robustness}
Nominal TES evaluation assumes perfect forecasts, but practical deployment does not. To test robustness under imperfect information, we perturb the cooling-demand forecast available to the controller while keeping environment transitions and scoring tied to the realized demand. Let $L_t$ denote the realized cooling demand and let the controller-visible forecast be
\begin{equation}
\hat L_t=L_t\left(1+\frac{e_t}{100}\right),
\end{equation}
where $e_t$ is a given forecast-bias rate at time step $t$. We evaluate a single time-varying underestimation scenario: during active cooling hours, $e_t$ ranges approximately from $-11.9\%$ to $-5.1\%$, while non-cooling hours have $0\%$ bias. The controller plans using $\hat L_t$, but TES state updates and unmet-demand penalties are computed from the ground-truth $L_t$. We focus on underestimation because it directly reduces the planned reserve margin, making it the more stringent direction for unmet-load risk in this benchmark. Earlier carbon-aware TES MPC work established load-forecast error as a relevant operational perturbation by evaluating both $+10\%$ and $-10\%$ forecast-bias cases \cite{miyata2020_dynamic_co2_mpc}. The carbon-intensity trajectory and all system parameters are held fixed so that forecast quality is the only manipulated factor. Relative to the nominal TES prompt, the LLM prompt in this stress test is changed only by adding a short warning that demand forecasts may be inaccurate, together with simple running statistics of forecast errors; the action interface and parser are unchanged (Appendix~\ref{app:prompts} lists the exact augmentation). For MPC, the same biased numerical forecast is provided without additional natural-language instruction.

For the post-RFT model, this scenario also tests the specificity of the training signal: as noted above, no aspect of forecast uncertainty was part of the verified reward. The stress test separates two questions: whether the planning behaviors reinforced by the verifier persist when the forecasts are inaccurate, and whether uncertainty-handling behavior emerges despite not being included in the verifier reward.

\begin{figure}[t]
\centering
\begin{minipage}[t]{0.47\textwidth}
\centering
\includegraphics[width=\textwidth]{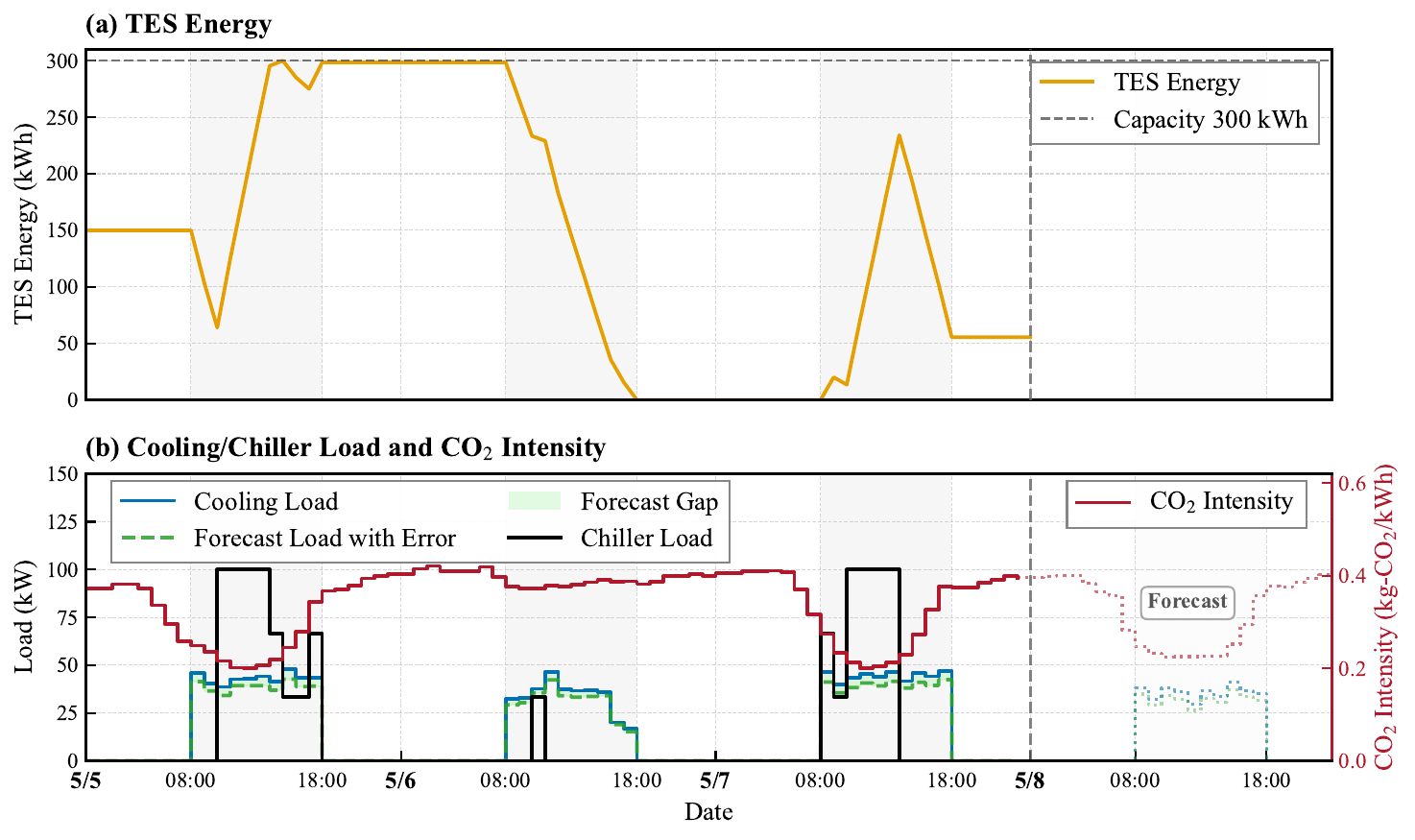}

{\small (a) MPC}
\end{minipage}\hfill
\begin{minipage}[t]{0.47\textwidth}
\centering
\includegraphics[width=\textwidth]{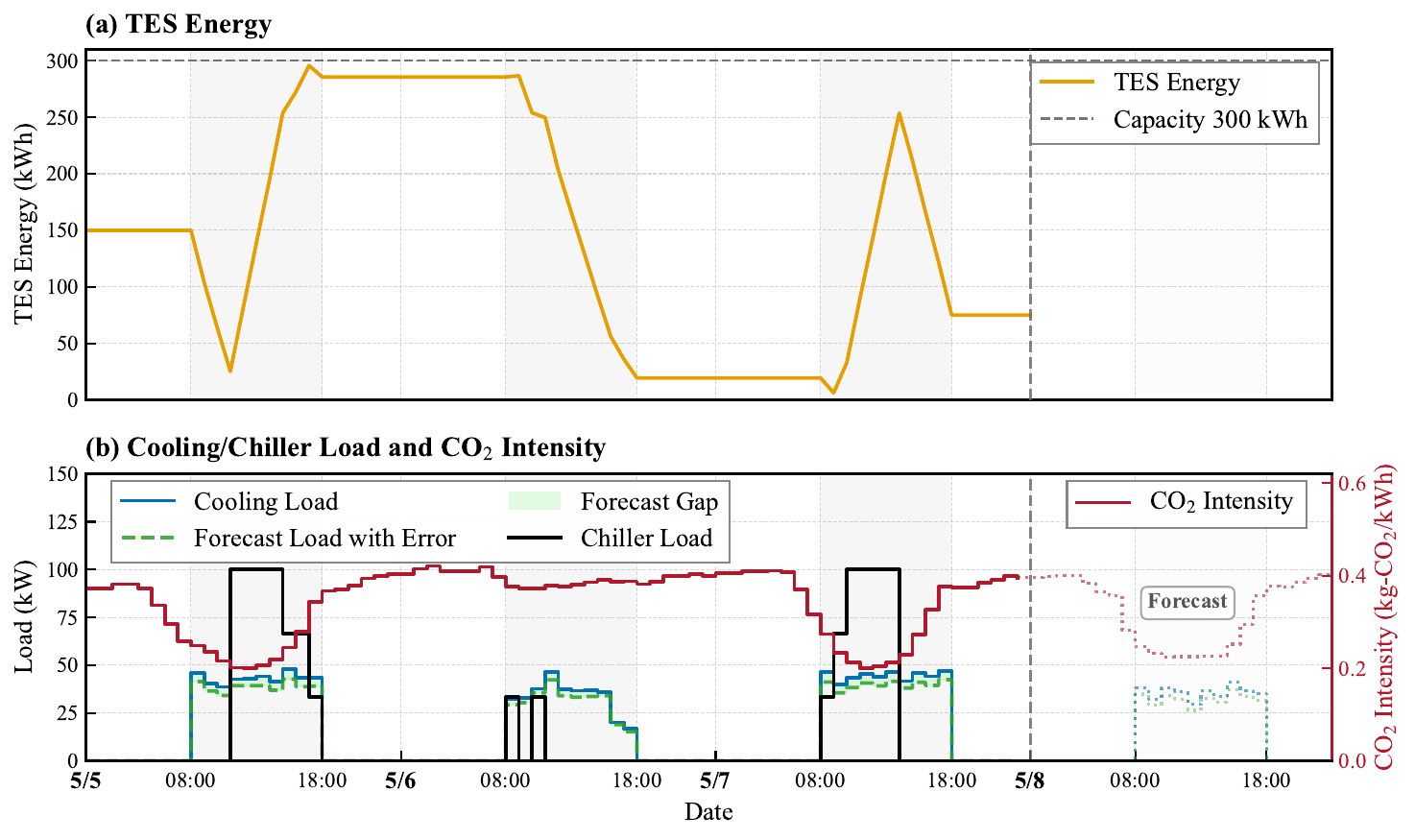}

{\small (b) GPT-5}
\end{minipage}

\vspace{0.8em}

\makebox[\textwidth][c]{%
\begin{minipage}[t]{0.47\textwidth}
\centering
\includegraphics[width=\textwidth]{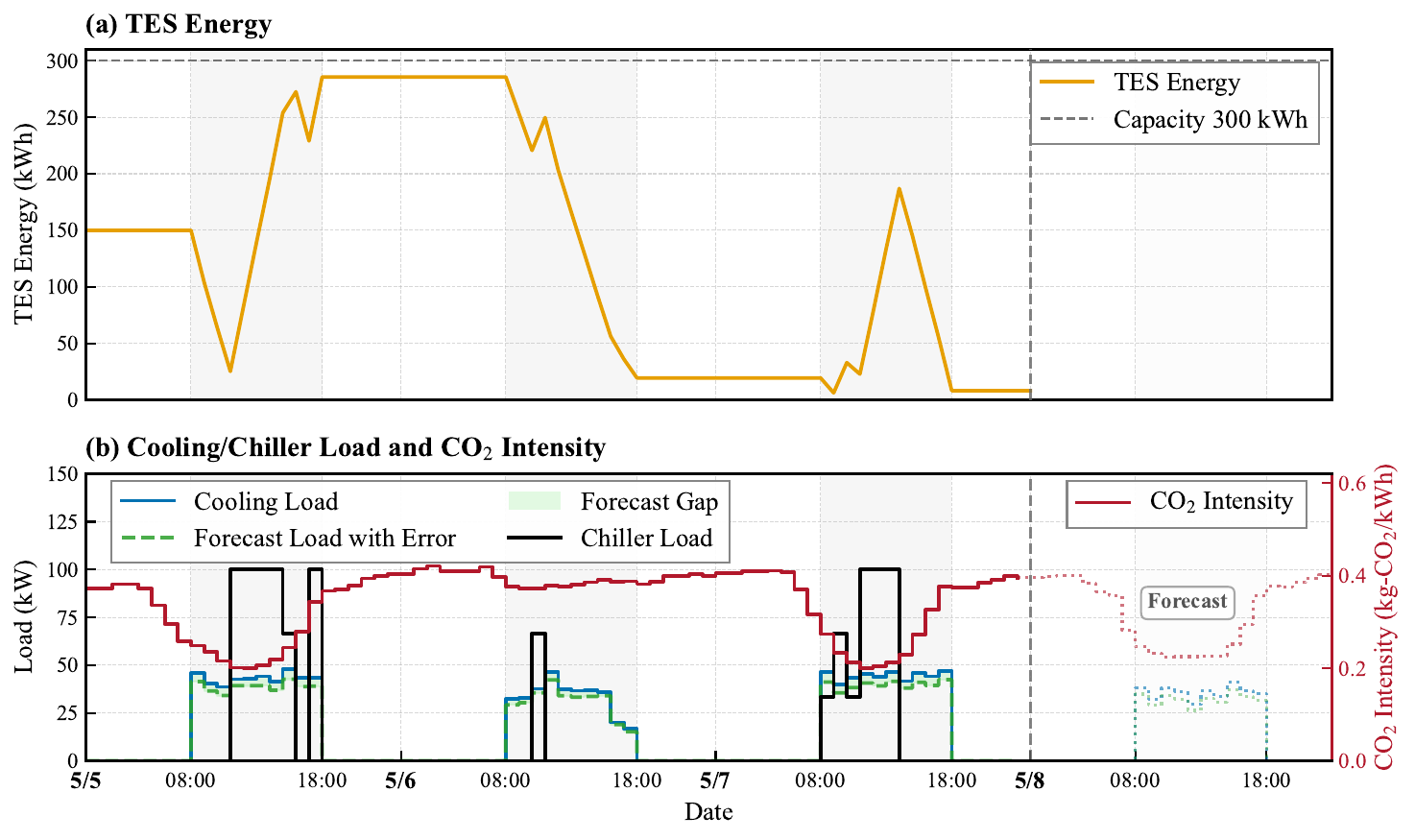}

{\small (c) gpt-oss-20b Post-RFT}
\end{minipage}
}
\caption{TES trajectories under forecast-error conditions for selected controllers. In each panel, the upper subplot presents the TES energy trajectory [kWh], while the lower subplot presents cooling load and chiller load [kW] together with grid CO$_2$ intensity [kg-CO$_2$/kWh].}
\label{fig:tes_forecast_error}
\end{figure}

Among the controllers that rely strongly on forecasts, MPC and GPT-5 provide contrasting examples of failure and success under this mismatch. As shown in Fig.~\ref{fig:tes_forecast_error}, MPC becomes vulnerable when demand is underestimated because it optimizes tightly against the forecasted trajectory and so tends to consume the available margin too aggressively; it was the only controller in our experiments to incur unmet cooling demand.

GPT-5, in contrast, shows substantially greater robustness. Its behavior under forecast error is not simply to become uniformly conservative; instead, it preserves a practical safety buffer while still exploiting lower-carbon hours when they are available. This is especially visible at 09:00 on May~7, 2025, the third day of the test episode, where GPT-5 selects a medium ASHP output level instead of full output. This choice balances the risk of demand underestimation against unnecessary overproduction.

The corresponding GPT-5 reasoning summary explicitly uses the forecast-bias information:
\begin{quote}
\small\itshape
``CO$_2$ is lower later (10--13) and much higher in late afternoon, and the load forecast is underestimating by $\sim$12\%, so we must avoid depleting the nearly empty TES while starting to build charge. Use medium ASHP now to cover possible higher load and modestly charge TES, reserving heavier charging for lower-CO$_2$ hours and discharging during the high-CO$_2$ period.''
\end{quote}

The post-RFT \texttt{gpt-oss-20b} shows a clear limit of the RFT effect: it does not account for forecast error. The generated traces treat the forecast as exact, emphasizing deterministic future loads and intensities even when the prompt provides a bias warning and running error statistics. At the same time, the trajectory in Fig.~\ref{fig:tes_forecast_error}(c) remains orderly. The planning behaviors reinforced by RFT (candidate comparison and feasibility checking against the stated constraints) continue to operate when the forecast is inaccurate, but no additional uncertainty-handling behavior emerges. Because the traces show no engagement with the stated uncertainty, we do not interpret this behavior as robust control; the absence of failure in this single underestimation scenario provides no basis for expecting similar behavior under other error patterns.

The stress test indicates that the effect of verifier-based RFT closely follows the reward design. Evidence from a different in-context-learning setting shows that larger models override semantic priors more readily when conflicting examples are provided in context \cite{ref:wei2023_icl_differently}. By analogy, the gap to GPT-5 may reflect a greater ability to treat the forecast-bias warning as a revised premise rather than retain the default assumption that the forecast is approximately correct. Uncertainty handling could therefore be improved in two ways: by using the stronger in-context learning capability of larger models at inference time, or by incorporating uncertainty into the verifier. For example, candidate actions could be scored against ensembles of perturbed demand trajectories so that the verified reward favors appropriate hedging.

\subsection{Generalization test settings}\label{subsec:generalization-settings}
Whereas the forecast-error test perturbs only the information given to the controller, the remaining two tests change the environment itself. As discussed in Section~\ref{subsec:rft-training-procedure}, we evaluate generalization through a condition shift within TES and a task transfer to battery scheduling. In both cases, the aim is not to maximize performance through additional adaptation but to test whether the same controller, prompt structure, and learned reasoning patterns remain useful when the numerical context changes.

The within-domain shift moves from the nominal Tokyo case to a Kyushu case with different external input trajectories and a larger storage system while keeping the TES decision interface unchanged. In the shifted case, the ASHP capacity is increased to 300~kW and TES capacity to 750~kWh, the reference peak cooling demand rises from about 48~kW to about 180~kW, and the carbon-intensity pattern becomes more strongly shaped by renewable curtailment, including near-zero midday values on some days. No additional fine-tuning is performed for this test. Table~\ref{tab:tes_shift_settings} summarizes the nominal and shifted condition settings.

\begin{table}[t]
\centering
\caption{Nominal and shifted TES condition settings used for the within-domain generalization test.}
\label{tab:tes_shift_settings}
\small
\begin{tabular}{lcc}
\toprule
Item & Tokyo nominal & Kyushu shift \\
\midrule
Reference peak cooling load & $\sim 48$ kW & $\sim 180$ kW \\
ASHP rated cooling capacity & 100 kW & 300 kW \\
TES capacity & 300 kWh & 750 kWh \\
Carbon-intensity range (reference) [kg-CO$_2$/kWh] & 0.20--0.421 & 0.00--0.351 \\
\bottomrule
\end{tabular}
\end{table}

To test cross-domain transfer, we replace the TES scheduling problem with a behind-the-meter photovoltaic (PV)-plus-battery scheduling problem. It shares the abstract structure of constrained storage scheduling but changes the physical semantics, time scale, and objective. Time is discretized at 30-min resolution over a single day. The battery has 10~kWh physical capacity, a usable state-of-charge (SoC) range of 1--9~kWh, maximum charge/discharge power of 3~kW, charge and discharge efficiencies of 0.90, and five discrete actions: discharge, hold, half-rate charge, full-rate charge, and PV-only charge. External inputs are household load, PV generation, and electricity price, all assumed perfectly known for that day. The objective is to minimize total grid-purchase cost, and the terminal value is set to zero because it is a single-day task. Table~\ref{tab:battery_environment} summarizes the battery task configuration. The TES-RFT model is evaluated on this battery task without battery-specific post-training.

\begin{table}[t]
\centering
\caption{Battery transfer-task settings.}
\label{tab:battery_environment}
\small
\begin{tabular}{@{}>{\raggedright\arraybackslash}p{0.30\textwidth}>{\raggedright\arraybackslash}p{0.36\textwidth}@{}}
\toprule
Item & Setting \\
\midrule
\mbox{Time step} & 30 min \\
\mbox{Evaluation horizon} & 1 d \\
\mbox{Physical battery capacity} & 10 kWh \\
\mbox{Usable SoC range} & 1--9 kWh \\
\mbox{Maximum charge/discharge power} & 3 kW \\
\mbox{Charge/discharge efficiency} & 0.90 / 0.90 \\
\mbox{Action set} & discharge, hold, half-rate charge, full-rate charge, PV-only charge \\
\mbox{External inputs} & household load, PV generation, electricity price \\
\mbox{Primary objective} & total grid-purchase cost \\
\mbox{Terminal value} & 0 \\
\bottomrule
\end{tabular}
\end{table}

\subsection{Generalization test results}\label{subsec:generalization-results}
\subsubsection{Cross-capacity and cross-region generalization}\label{subsec:capacity-region}
The within-domain generalization test shifts the numerical regime from the Tokyo nominal case to the Kyushu case described in Section~\ref{subsec:generalization-settings}. If the post-RFT controller had merely memorized the nominal Tokyo scale, its performance would be expected to degrade sharply under this shift.

\begin{figure}[t]
\centering
\includegraphics[width=\textwidth]{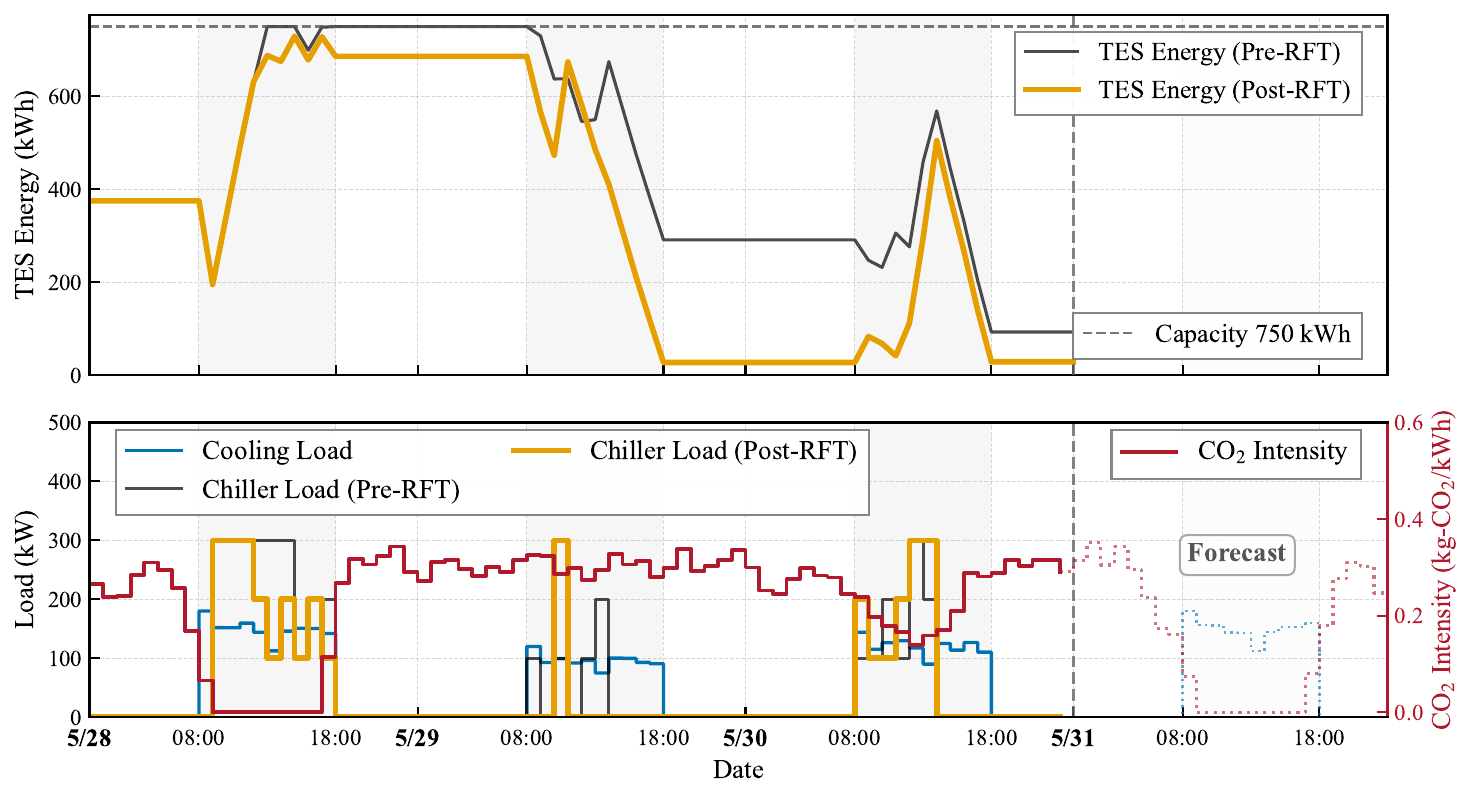}
\caption{Pre/Post-RFT comparison of \texttt{gpt-oss-20b} in the Kyushu TES condition-shift test.}
\label{fig:kyushu_compare_main}
\end{figure}

Table~\ref{tab:kyushu_results} and Fig.~\ref{fig:kyushu_compare_main} show that the results do not exhibit such a sharp degradation. GPT-5 again remains close to the DP reference, achieving 75.5~kg-CO$_2$ versus 74.5~kg-CO$_2$ for DP. For the RFT comparison, the post-RFT \texttt{gpt-oss-20b} reaches 76.4~kg-CO$_2$, while the pre-RFT model remains substantially higher at 85.1~kg-CO$_2$, so most of the RFT-induced gain survives even though both the equipment scale and the external input trajectories have changed markedly. For reference, operating without storage on the same Kyushu episode yields 141.0~kg-CO$_2$, so the post-RFT model cuts emissions by about 46\% relative to the no-storage baseline even under this unseen condition.

The generated traces are consistent with this interpretation. For example, a representative Pre-RFT Kyushu trace shows a direct jump from the presence of zero-carbon hours to immediate charging:
\begin{quote}
\small\itshape
``But still better to charge at zero CO2 to avoid future 0.274 usage. So action [3]''
\end{quote}
By contrast, a representative Post-RFT trace shows explicit feasibility re-checking before committing to the action:
\begin{quote}
\small\itshape
``TES currently 729, +149=878 >750 capacity \ldots\\
We should not exceed capacity.\\
Let's double-check constraints \ldots''
\end{quote}
This suggests that what carries over to the Kyushu case is not a fixed Tokyo-specific action schedule, but a reusable verifier-aligned habit of checking feasibility before exploiting low-carbon opportunities.

\begin{table}[t]
\centering
\caption{TES condition-shift results for the Kyushu generalization test.}
\label{tab:kyushu_results}
\small
\begin{tabular}{l@{\hspace{2.5em}}r}
\toprule
Controller & Total emissions [kg-CO$_2$] \\
\midrule
Baseline (No Thermal Storage) & 141.0 \\
\mbox{gpt-oss-20b} Pre-RFT & 85.1 \\
\mbox{gpt-oss-20b} Post-RFT & 76.4 \\
GPT-5 & 75.5 \\
DP & 74.5 \\
\bottomrule
\end{tabular}
\end{table}

\subsubsection{Transfer to the battery task}\label{subsec:battery-transfer}
The battery experiment tests transfer to a related storage-scheduling task with different physical semantics and a different objective. The question here is which parts of the learned TES reasoning process remain useful without battery-specific post-training.

Table~\ref{tab:battery_transfer_results} shows a clear but incomplete transfer pattern. GPT-5 again reaches near-optimal performance, with a total cost of 153~yen versus 152~yen for DP. The TES-RFT \texttt{gpt-oss-20b} improves substantially over its pre-RFT version, reducing cost from 218 to 194~yen, but it remains well above the optimum. Some planning patterns learned on TES transfer, but not enough to match the strongest controllers on the battery task.

\begin{figure}[t]
\centering
\begin{minipage}[t]{0.47\textwidth}
\centering
\includegraphics[width=\textwidth]{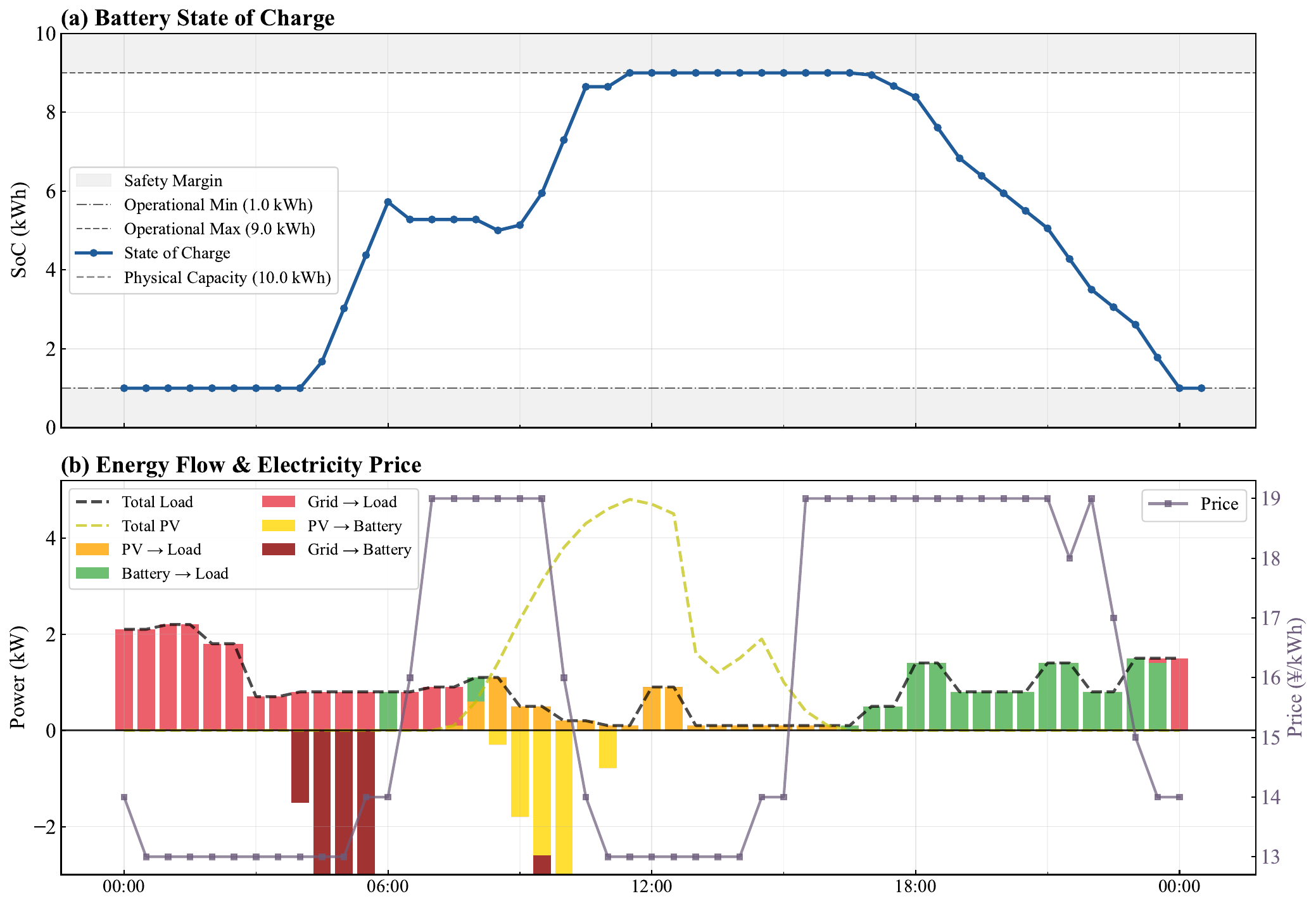}

{\small (a) gpt-oss-20b Pre-RFT}
\end{minipage}\hfill
\begin{minipage}[t]{0.47\textwidth}
\centering
\includegraphics[width=\textwidth]{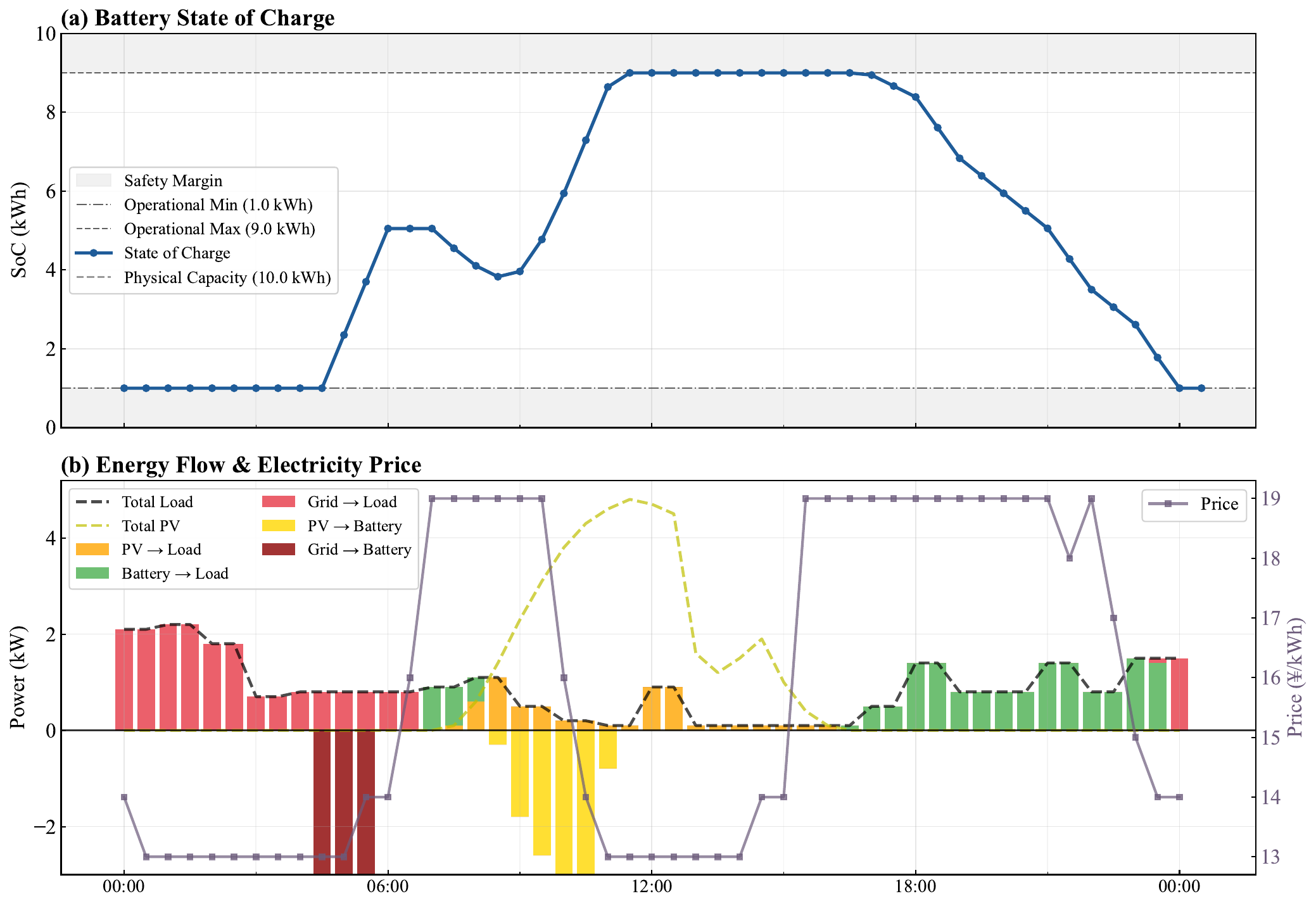}

{\small (b) gpt-oss-20b Post-RFT}
\end{minipage}

\vspace{0.8em}

\begin{minipage}[t]{0.47\textwidth}
\centering
\includegraphics[width=\textwidth]{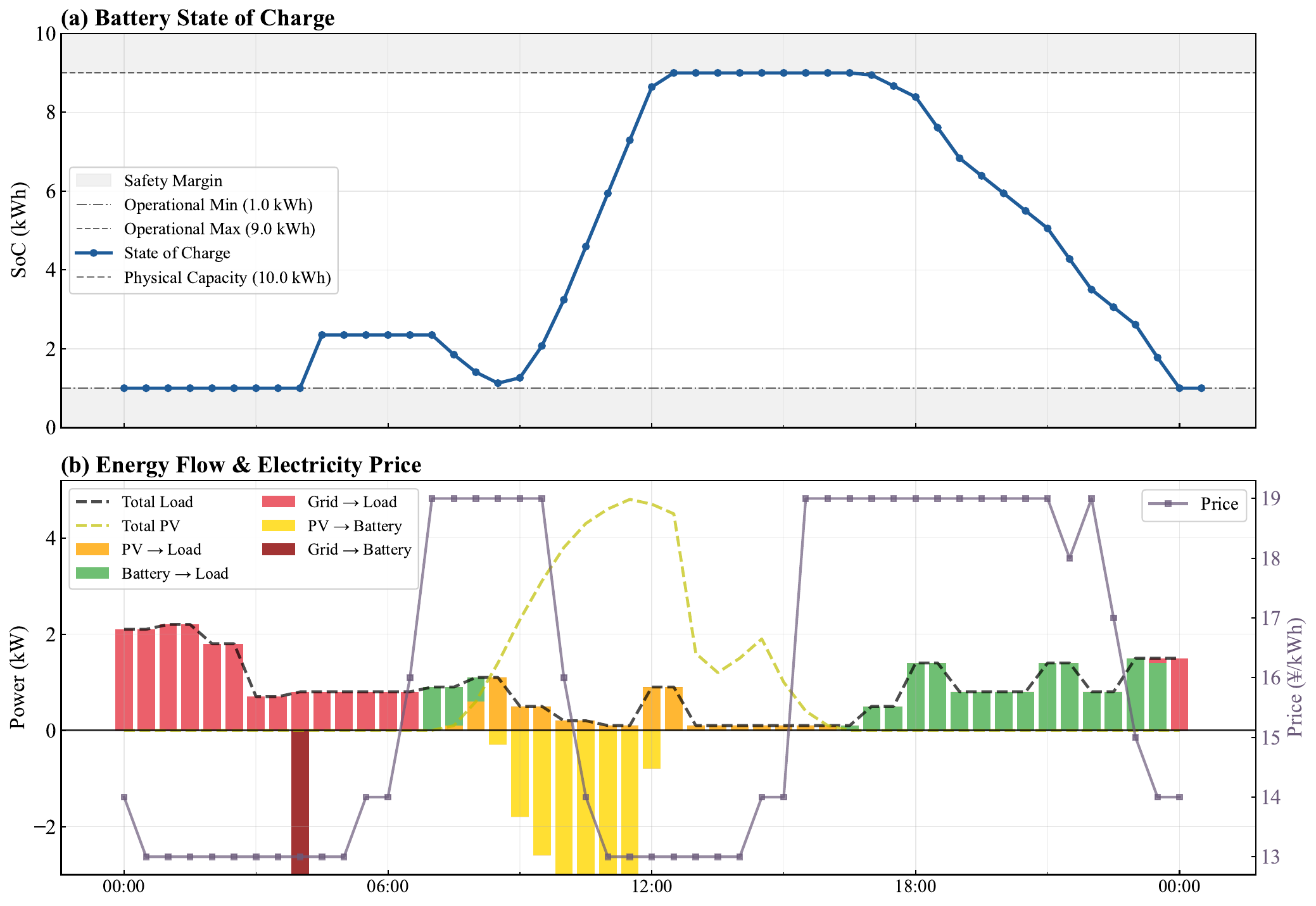}

{\small (c) GPT-5}
\end{minipage}\hfill
\begin{minipage}[t]{0.47\textwidth}
\centering
\includegraphics[width=\textwidth]{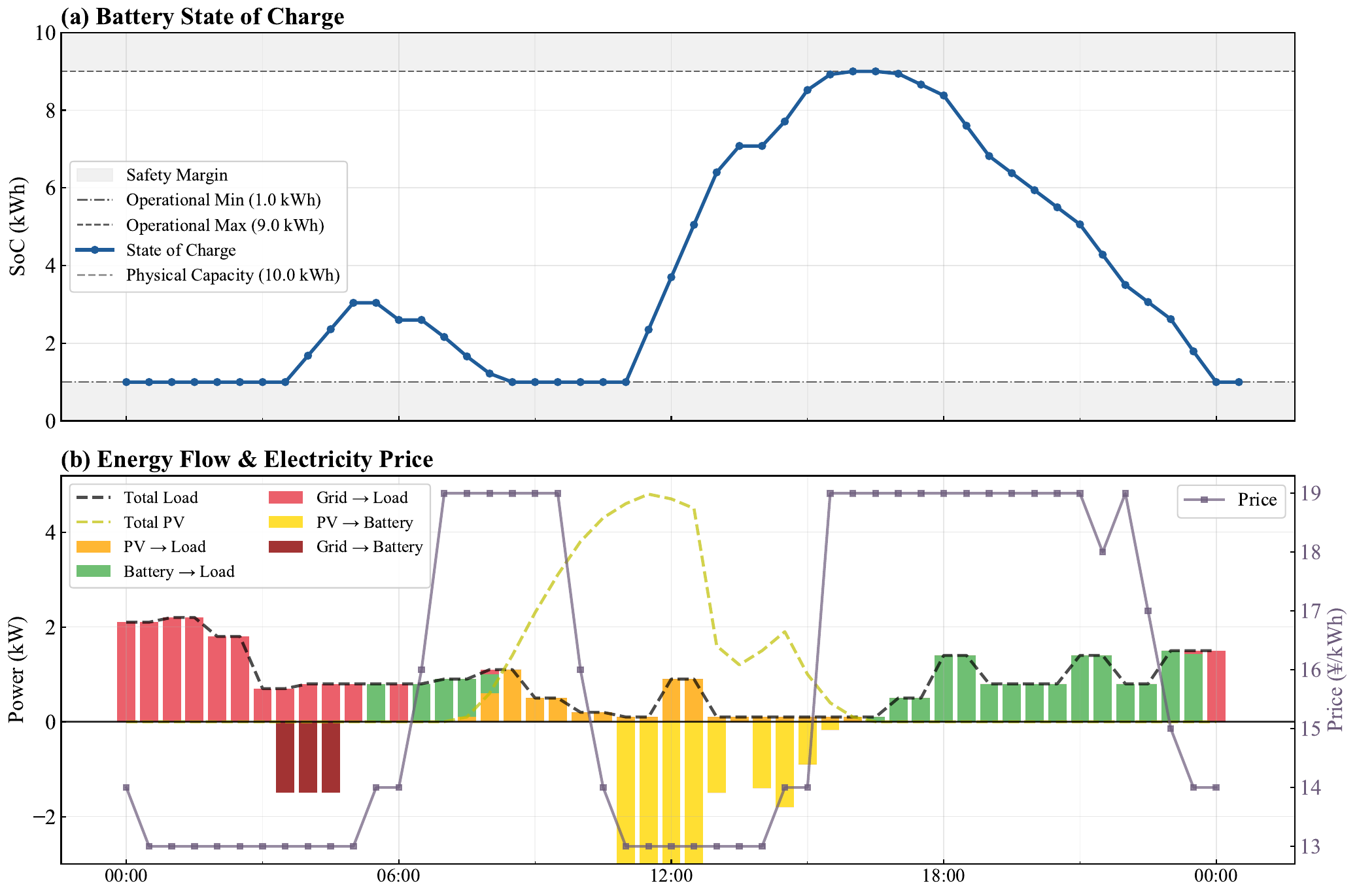}

{\small (d) DP}
\end{minipage}
\caption{Battery trajectories for selected controllers in the transfer task. In each panel, the upper subplot presents the battery state trajectory [kWh], while the lower subplot presents load [kW], PV generation [kW], and electricity price [yen/kWh].}
\label{fig:battery_transfer_main}
\end{figure}

Figure~\ref{fig:battery_transfer_main} and the associated generated traces clarify the source of this remaining gap. For example, after RFT the model more often externalizes a multi-step planning loop:
\begin{quote}
\small\itshape
``Which charging action? \ldots We want to charge as much as possible before high price periods. \ldots Let's analyze.''
\end{quote}
However, this improved process is still paired with an incomplete battery-specific strategy. As shown in Fig.~\ref{fig:battery_transfer_main}(b), the post-RFT model raises the battery state before PV generation becomes available, whereas the DP trajectory in Fig.~\ref{fig:battery_transfer_main}(d) preserves more headroom for the PV-rich midday period. The trace excerpt is consistent with this behavior: it focuses on charging before high-price periods, but does not explicitly reason about the onset of PV generation, PV-only charging, or the opportunity cost of filling the battery before zero-marginal-cost PV energy becomes available.

\begin{table}[t]
\centering
\caption{Battery-task results for zero-shot transfer from TES-trained reasoning.}
\label{tab:battery_transfer_results}
\small
\begin{tabular}{l@{\hspace{2.5em}}r}
\toprule
Controller & Total cost [yen] \\
\midrule
\mbox{gpt-oss-20b} Pre-RFT & 218 \\
\mbox{gpt-oss-20b} Post-RFT & 194 \\
GPT-5 & 153 \\
DP & 152 \\
\bottomrule
\end{tabular}
\end{table}

\subsection{What persists and what does not}\label{subsec:transfer-limits}
Ordered by distance from the verified training condition, the three tests give a consistent picture. Under corrupted forecast information, the closest departure, the reinforced planning behaviors remain intact while uncertainty handling, which the verifier never scored, does not appear (Section~\ref{subsec:robustness}).

The Kyushu and battery tests represent larger departures from the training condition. Together, they suggest that TES-specific RFT shapes reasoning at an intermediate level of abstraction. The most transferable elements are those already visible in the generated traces: explicit look-ahead, comparison of candidate actions, and final feasibility checking. These elements are sufficient to preserve strong performance when the target problem retains the same state and action semantics but changes the numerical regime.

The task-specific strategy transfers less completely. The battery task changes not only the numerical regime but also the available opportunities, because PV is a time-limited energy source with zero marginal cost. As shown in Section~\ref{subsec:battery-transfer}, the post-RFT model retains deliberate candidate evaluation but does not reliably account for the PV-specific timing and headroom trade-off. Thus, RFT can stabilize a reusable reasoning pattern for constrained storage control, but it does not produce a fully domain-agnostic controller. Transfer is strongest when the target problem preserves the physical semantics and changes only the numerical regime. Across all three tests, RFT strengthens the constraint-aware planning process scored by the verifier, but does not add behavior that the verifier never rewards. This is the verdict on the generalization component of the hypothesis in Section~\ref{subsec:gap-objectives}: the reusable procedure carries across operating scales, but only partially across physical semantics.

\section{Discussion}\label{sec:discussion}
\subsection{Interpreting the effect of verifier-based RFT}\label{subsec:why-it-works}
Taken together, the nominal results support a more specific interpretation than the broad claim that an LLM can control TES. After training on only 30 prompts, verifier-based RFT brings the open-weight model to within 0.4~kg-CO$_2$ of the DP reference on the held-out May test episode. Yet the best-of-10 diagnostic shows that a near-optimal trajectory can already be sampled from the Pre-RFT model. Together, these findings suggest that post-training does not need to create a new scheduling strategy. Instead, it makes the base model more likely to use an existing multi-step decision procedure consistently.

The action-value reward design provides a plausible mechanism for this effect. Rather than reducing each training state to a single target action, the DP verifier retains the relative long-horizon value of every admissible action. Training can then distinguish poor, near-optimal, and optimal choices while reusing the same small set of verified states across many rollout groups and optimizer updates. In this setting, the computational cost of DP is paid offline and amortized through post-training, while the pretrained model supplies a prior over how to compare alternatives, anticipate delayed consequences, and check feasibility.

The behavioral evidence reinforces this interpretation. After RFT, action agreement with DP rises, early charging errors become less frequent, and the generated traces more reliably exhibit candidate comparison, forward simulation, and final constraint checking. These changes show that RFT makes effective decision behavior more consistent when long-horizon comparisons matter.

The present result should not be taken as an upper bound on what verifier-based training can produce. At the scale studied here---a 20B-parameter open-weight model, LoRA-only adaptation, and 30 training prompts---RFT makes existing planning behavior more reliable. Larger-scale reinforcement learning has elicited reasoning behaviors that were weak or absent before post-training \cite{ref:deepseekr1_2025}, and analyses of RLVR indicate that group-relative updates can amplify initially rare successful outputs \cite{mroueh2025rlvr}. Whether larger models, longer verifier-based training, or verifiable-reward objectives applied earlier in training can produce qualitatively new control strategies remains open.

\subsection{How the verifier shapes generalization}\label{subsec:verifier-generalization}
Across the robustness and transfer tests, generalization appears to follow the decision structure captured by the verifier reward. Candidate comparison, look-ahead, and feasibility checking remain useful when the numerical regime changes from the Tokyo case to the larger Kyushu system, where the state and action semantics are preserved. These behaviors also appear in the battery task, but they are not enough to produce the PV-specific strategy needed to preserve headroom for zero-marginal-cost charging. Under forecast error, the reinforced deterministic planning procedure remains orderly, yet the model does not learn to hedge because the verifier never rewards uncertainty-aware behavior.

The learned procedure falls between memorization and a domain-independent policy. It is more general than a memorized action schedule: the gain persists on a held-out TES episode and under a substantial change in equipment scale and regional inputs. It is less general than a domain-independent policy: when the physical meaning of the state, actions, and future opportunities changes, only part of the benefit transfers. Generalization therefore depends less on how much the numbers change than on whether the target setting preserves the decision structure captured by the verifier.

These findings also clarify when transfer across buildings is plausible. A controller adapted in one TES setting may remain useful when capacities, loads, and carbon signals change but the underlying storage semantics and interface remain stable. A new equipment domain is more likely to require a domain-specific verifier and additional adaptation. For the battery task, such a verifier would need to reward PV-aware headroom management; for forecast-robust TES control, it would need to evaluate candidate actions across multiple demand trajectories with simulated forecast errors. At the model scale studied here, either capability would likely require an explicit verifier-based training signal rather than inference-time prompting alone.

\subsection{Implications for energy-control deployment}\label{subsec:framework-outlook}
Verifier-based RFT does not remove model-based computation; it moves it offline. In this study, a system model and DP are used during training to assign long-horizon values to candidate actions. These values are computed offline and reused across many rollouts and optimizer updates. After training, the model selects actions from a structured observation and prompt, and an action parser converts its output into a control command; no DP or other online optimization is required during deployment. The transfer results further suggest the potential for zero-shot deployment to new buildings: when the TES semantics and observation--action interface are preserved, the same adapted model may be used without site-specific retraining.

The framework itself could extend beyond TES to the control of whole-building energy systems, including variable-air-volume (VAV) systems and air-handling units (AHUs). Such an extension would require a control problem with explicit constraints, a solver or simulator that can score candidate actions, and a stable observation--action interface.

Language-based controllers can receive equipment constraints, operating rules, and forecast caveats through one interface, though access does not ensure effective use (Section~\ref{subsec:robustness}). GPT-5 provides a frontier reference without task-specific training, whereas the open-weight experiments test verifier-based adaptation in a self-hosted model. For hourly control across many buildings, self-hosting may offer data-local or edge operation, trace-level inspection, lower marginal inference costs, and less dependence on network connectivity and latency. These trade-offs keep task-adapted open-weight controllers relevant without making either model universally preferable.

Although not evaluated here, the same verifier could support verifier-in-the-loop self-improvement of an agentic harness around a fixed model. During offline development, DP-derived action values could be used to compare alternative contexts, retrieved examples, memory contents, and tool-use policies based on the control decisions they produce. The resulting feedback could guide context curation, memory updates, and tool selection or sequencing, shifting adaptation from repeated weight updates toward system-level harness optimization. The forecast-error traces also provide a useful diagnostic: the stated reasoning does not engage with uncertainty. Compared with DQN's numerical actions and value estimates, the open-weight model's traces make its stated assumptions and candidate comparisons directly inspectable; they are also more detailed than GPT-5's brief reasoning summaries.

\subsection{Limitations}\label{subsec:limitations}
These results come from the simplified simulation benchmark adopted in the Introduction. The TES model assumes a constant COP, discrete actions, and a three-day horizon, and it omits storage heat losses, thermal stratification, and mixing dynamics. These simplifications make exact DP tractable, allowing the same action values to serve as verifier rewards and as the optimal reference. Forecasts are also assumed to be accurate except in the dedicated forecast-error test. The results show that the reward-construction and post-training method works in this controlled setting; they do not establish deployment-ready TES control. Testing the method under variable COP, tank heat losses and mixing, season-long horizons, diverse load and weather conditions, sensing errors, missing data, and safety requirements will require higher-fidelity system models.

The robustness and generalization tests cover three qualitatively different departures from the training condition, with one case in each direction: forecast error, a shifted numerical regime, and changed physical semantics. This design reveals what persists and what does not, but breadth within each direction (other error patterns, operating scales and regions, and storage tasks) remains to be tested.

Generated traces can contain calculation errors or unsupported statements. For example, one battery-task trace stated:
\begin{quote}
\small\itshape
``At 09:30 PV 3.10, load 0.50, surplus 2.6 kWh. So 1.5 kWh charge uses\\
PV surplus; grid not needed.''
\end{quote}
Here, the trace treats a power surplus of 2.6~kW as 2.6~kWh of energy and ignores the 30-min control interval; the available surplus energy is only 1.3~kWh before charging losses. This is more than a minor arithmetic error: it shows that the model does not consistently apply the basic relation between power, energy, and time. Better control performance therefore does not guarantee reliable physical reasoning in the generated traces. Future work should examine more rigorously what these traces can and cannot explain.

The baseline comparison is intended as a controlled evaluation under a common simulator and interface, not as a definitive ranking of controller families.

The main challenge in extending the method is the verifier. Exact backward DP works for the compact discretized problem studied here, but it becomes difficult for higher-fidelity TES, VAV, and AHU systems with nonlinear dynamics, interacting devices, longer horizons, and large or continuous state and action spaces. RL-based Q-value approximators, critic networks, or other surrogate value models could provide approximate rewards, but those rewards would no longer be exactly verifiable. How accurate such a verifier must be for verifier-based training to remain reliable is an open question.

\section{Conclusion}\label{sec:conclusion}
This study examined whether the objective of a constrained energy-control problem can be converted into a verifiable learning signal for post-training an open-weight reasoning model. TES scheduling was formulated as an RLVR task in which exact DP was used to compute action values for every admissible action at each verified state. These values were transformed into dense verifier rewards and used to fine-tune the model as an upper-level TES scheduler.

Using only 30 training prompts, verifier-based RFT reduced the open-weight model's total emissions from 70.5 to 61.2~kg-CO$_2$ on a temporally held-out test episode, bringing its performance close to the DP optimum of 60.8~kg-CO$_2$. The improvement was accompanied by more consistent observable planning behavior, including explicit comparison of candidate actions, anticipation of future consequences, and feasibility checking. The best-of-10 diagnostic further showed that a near-optimal action sequence was already accessible within the Pre-RFT model's sampling distribution. Together, these results suggest that RFT primarily increased the reliability with which the model selected effective actions, rather than inducing a qualitatively new scheduling strategy.

The robustness and generalization tests further showed that the effect of RFT follows the reward design. The planning behaviors strengthened by RFT remained visible under demand-forecast error, and the performance gain persisted without additional adaptation under an unseen TES condition with different equipment capacities and regional carbon-intensity patterns. However, uncertainty-aware hedging---which was not represented in the verifier reward---did not emerge. Transfer to the battery-scheduling task was also only partial. General planning behaviors remained observable, but changes in physical semantics, temporal resolution, and future opportunities limited the performance gain. The transferable outcome should therefore be interpreted as a verifier-aligned planning procedure rather than a domain-independent control policy.

The cross-model comparison also suggests that inference-time reasoning is important for this scheduling problem. GPT-5 nearly matched DP and MPC without task-specific training and was the only controller that adapted its decisions to the stated forecast bias, whereas GPT-4o performed worse than the no-storage baseline. Although this comparison does not isolate all differences in model architecture and training, it indicates that fluent instruction following alone may be insufficient for reliable long-horizon storage scheduling under operational constraints.

Overall, the results demonstrate the feasibility of deriving verifiable post-training signals directly from a building-energy control objective. The present study nevertheless remains a proof of concept based on a simplified simulation. Future work should evaluate the framework using higher-fidelity models of TES, VAV, and AHU systems; incorporate forecast uncertainty and safety constraints into the verifier; and develop approximate value-based verifiers for problems in which exact DP is computationally intractable. Reliable approximate verifiers would also open the path from single components toward whole-building and, eventually, district- or city-scale energy management. Future work could also investigate verifier-in-the-loop self-improvement of agentic control harnesses during offline development, using DP- or critic-based feedback to curate context and memory and refine tool-use policies. Verifier-based post-training offers a practical mechanism for transferring optimization knowledge into reasoning-model controllers, and this study lays the groundwork for RLVR in energy management.

\section*{CRediT authorship contribution statement}
\textbf{Takumi Shioda:} Conceptualization, Methodology, Software, Validation, Formal analysis, Investigation, Data curation, Writing -- original draft, Writing -- review \& editing, Visualization. \textbf{Kohei Terashima:} Writing -- review \& editing. \textbf{Tatsuo Nagai:} Supervision, Writing -- review \& editing.

\section*{Funding}
This research did not receive any specific grant from funding agencies in the public, commercial, or not-for-profit sectors.

\section*{Declaration of competing interests}
The authors declare that they have no known competing financial interests or personal relationships that could have appeared to influence the work reported in this paper.

\section*{Data availability}
Data will be made available on request.

\section*{Declaration of generative AI and AI-assisted technologies in the manuscript preparation process}
During the preparation of this work, the authors used OpenAI ChatGPT to improve the English language and readability of the manuscript. After using this tool/service, the authors reviewed and edited the content as needed and take full responsibility for the content of the published article.

\thispagestyle{fancy}

\clearpage
\appendix
\raggedbottom
\titleformat{\section}{\color{ink}\Large\bfseries}{Appendix~\thesection}{0.85em}{}
\counterwithin{table}{section}
\counterwithin{figure}{section}

\section{Baseline Hyperparameters}\label{app:baseline-hparams}
This appendix lists the main settings and hyperparameters used for the DP, MPC, and RL baselines in the TES experiments. The same system dynamics, action set, objective, unmet-demand penalty, and terminal-value definition were used across the baselines. The DP uses 2{,}000 bins for TES-energy discretization. Tables~\ref{tab:mpc_de_hparams}--\ref{tab:ppo_hparams} summarize the MPC, DQN, and PPO configurations, respectively.

\begin{table}[H]
\centering
\caption{Main hyperparameters for MPC with differential evolution.}
\label{tab:mpc_de_hparams}
\small
\begin{tabular}{@{}>{\raggedright\arraybackslash}p{0.30\textwidth}>{\raggedright\arraybackslash}p{0.36\textwidth}@{}}
\toprule
Item & Setting \\
\midrule
Prediction horizon & 36 h \\
Decision variables & Discrete action sequence over the horizon \\
Population size parameter & 10 \\
Mutation range & $(0.5,1.0)$ \\
Recombination rate & 0.7 \\
Maximum iterations & 120 \\
Convergence tolerance & $10^{-3}$ \\
\bottomrule
\end{tabular}
\end{table}

\begin{table}[H]
\centering
\caption{Main hyperparameters for DQN.}
\label{tab:dqn_hparams}
\small
\begin{tabular}{@{}>{\raggedright\arraybackslash}p{0.30\textwidth}>{\raggedright\arraybackslash}p{0.36\textwidth}@{}}
\toprule
Item & Setting \\
\midrule
Algorithm & Double DQN with replay buffer and target network \\
Training period & April--June 2024 \\
Training episodes & 30,000 \\
Optimizer & Adam \\
Learning rate & $3\times10^{-4}$ \\
Activation function & ReLU \\
Discount factor $\gamma$ & 0.995 \\
Target-update coefficient & 0.01 \\
Replay-buffer size & 70,000 \\
Batch size & 256 \\
Learning-start step & 100 \\
Gradient updates per step & 2 \\
$\epsilon$ schedule & 1.0 to 0.05 over 50,000 steps \\
Hidden layers & $(256,256)$ \\
Weight decay & $10^{-4}$ \\
\bottomrule
\end{tabular}
\end{table}

\begin{table}[H]
\centering
\caption{Main hyperparameters for PPO.}
\label{tab:ppo_hparams}
\small
\begin{tabular}{@{}>{\raggedright\arraybackslash}p{0.30\textwidth}>{\raggedright\arraybackslash}p{0.36\textwidth}@{}}
\toprule
Item & Setting \\
\midrule
Training episodes & 30,000 \\
Training period & April--June 2024 \\
Optimizer & Adam \\
Actor/critic learning rates & $3\times10^{-4}$ / $3\times10^{-4}$ \\
Activation function & ReLU \\
Discount factor $\gamma$ & 0.995 \\
GAE parameter $\lambda$ & 0.95 \\
Clipping ratio & 0.2 \\
Entropy coefficient & 0.01 \\
Value-loss coefficient & 0.5 \\
Update epochs & 2 \\
Batch size & 256 \\
Rollout length & Learning-window terminal step \\
Actor/critic hidden layers & $(256,256)$ / $(256,256)$ \\
\bottomrule
\end{tabular}
\end{table}

\section{Prompt Templates and Output Constraints}\label{app:prompts}
This appendix records the fixed prompt interface used to evaluate the LLM controllers. The exact numerical values change by time step, but the role decomposition and output constraints remain fixed so that differences in behavior can be attributed to the controller, not to prompt redesign.

\subsection{TES prompt structure}
The TES controller uses the following fixed prompt template. For readability, the appendix shows descriptive placeholders in angle brackets rather than the literal variable-style placeholders used in the implementation.

\promptrole{System}
\begin{lstlisting}[style=promptblock]
You are an optimization agent that supervises a thermal energy storage (TES) plant. Your job is to minimize cumulative CO2 emissions over the full simulation horizon by planning TES charge and discharge decisions.
\end{lstlisting}

\promptrole{Developer}
\begin{lstlisting}[style=promptblock]
Developer instructions: respond using ASCII characters only. Return a single line formatted exactly as [n], where n is an integer in {0, 1, 2, 3}. Do not include additional text, explanations, markdown, or keys.
\end{lstlisting}

\promptrole{User}
\begin{lstlisting}[style=promptblock]
Objective:
- Minimize total CO2 emissions = electricity consumption times time-varying CO2 intensity over the horizon.

Current context:
- Current time [h]: <current time>
- Current TES energy [kWh]: <current TES energy>

Forecast data:
<forecast table>

System parameters:
ASHP rated capacity [kW]: <ASHP rated capacity>
ASHP base COP [-]: <ASHP base COP>
TES capacity [kWh]: <TES capacity>

Action space for the next hour:
0 -> ASHP output ratio = 0.00 (ASHP off; rely on TES if demand exists)
1 -> ASHP output ratio ~= 0.33 (low output; TES covers most of the remaining demand)
2 -> ASHP output ratio ~= 0.67 (medium output; TES supplements when load exceeds this level)
3 -> ASHP output ratio = 1.00 (full output; any surplus charges TES if capacity remains)

Operational notes:
- TES discharges automatically when load exceeds the scheduled ASHP output and energy is available.
- TES charges automatically when ASHP output exceeds the load and free capacity exists.
- Always maintain TES energy within the allowed bounds (<minimum TES energy> to <maximum TES energy>); violating these constraints incurs a large penalty.

Decision requirements:
- Optimize with a full-horizon perspective rather than a greedy step.
- Keep TES utilization efficient; avoid unnecessary saturation or depletion.
- Prioritize emission reductions even if it requires near-term energy use.
- Consider pre-charging during low-carbon periods and discharging during high-carbon periods while respecting TES energy limits.
- Consider next-day information when making decisions.

Return format:
- Output a single token formatted as [0], [1], [2], or [3].
\end{lstlisting}

This template is kept fixed between nominal evaluation and TES-specific RFT.

\subsection{GPT-5 JSON output variant for TES}
GPT-5 uses the same system prompt and the same user-side control context, but the output constraint differs because a brief rationale summary is logged together with the action. The developer-level output instruction is:

\begin{lstlisting}[style=promptblock]
Return a single JSON object on one line exactly as {"thought_process": "<brief reasoning>", "action": [<integer 0-3>]}. thought_process must be a concise summary (1-3 sentences).
\end{lstlisting}

Under this variant, only the integer inside the \texttt{action} field is executed by the simulator. The \texttt{thought\_process} field name is reproduced here because it is the literal output key used in the experiment, but in the analysis it is treated only as an externalized reasoning summary for post hoc comparison. This distinction matters because the open-weight model evidence mainly comes from longer rationale-bearing completions collected during RFT evaluation, whereas GPT-5 reasoning evidence is available only through this short summary-style field rather than through a full raw reasoning process.

\subsection{Battery prompt structure}
The cross-domain battery task preserves the same role decomposition but replaces the physical semantics and action set.

\promptrole{System}
\begin{lstlisting}[style=promptblock]
You are an optimization agent that schedules a behind-the-meter battery and PV system. Your goal is to minimize total grid-electricity cost while always keeping the load supplied over the full 30-minute resolution horizon. Think several steps ahead and respect all physical limits.
\end{lstlisting}

\promptrole{Developer}
\begin{lstlisting}[style=promptblock]
Developer instructions: respond using ASCII characters only. Return a single token formatted exactly as [n], where n is an integer in {0, 1, 2, 3, 4}. Do not emit any commentary, code fences, or additional symbols.
\end{lstlisting}

\promptrole{User}
\begin{lstlisting}[style=promptblock]
Objective:
- Minimize cumulative grid-purchase cost (grid imports times yen/kWh) while satisfying every 30-minute load.

Current SoC: <current battery energy> kWh (<current SoC percentage>% of the <battery capacity> kWh capacity).
Decision applies to interval starting <current time> (step <current step>/<total number of steps>).

Battery specs:
- Max charge/discharge power: <maximum battery power> kW (=> <maximum transferable energy> kWh per 30 min).
- Charge efficiency: <charge efficiency>; discharge efficiency: <discharge efficiency>; round-trip ~= <round-trip efficiency>.
- Actions obey automatic PV priority: PV -> load -> (if allowed) PV -> charge -> grid covers the rest.

Action space (return [n] only):
0 -> DISCHARGE: supply up to <maximum battery power> kW (<maximum transferable energy> kWh/interval) from the battery to offset load; grid only covers any leftover load.
1 -> HOLD: keep the battery idle; PV handles load first, grid supplies any deficit; no charging.
2 -> CHARGE_HALF: request ~<half charging power> kW (<half transferable energy> kWh/interval) charging; PV surplus is used first, grid supplies the rest.
3 -> CHARGE_FULL: request <maximum battery power> kW (<maximum transferable energy> kWh/interval) charging; PV surplus first, then grid.
4 -> PV_ONLY_CHARGE: charge only with PV surplus (up to <maximum battery power> kW) and never draw grid energy for charging; grid may still serve unmet load.

Operating policy reminders:
- Never violate SoC bounds [0, capacity].
- PV production always serves the load first; only surplus may charge the battery.
- Charging commands may draw from the grid after PV surplus unless the action explicitly forbids it.
- Grid purchases cost imported energy times electricity price. Minimize this total across all steps.
- Optimize holistically across the entire forecast - avoid greedy myopic moves.

Full forecast (all known predictions; '*' marks the current step):
timestamp, load, PV, electricity price, marker
<time 0>, <load 0>, <PV 0>, <price 0>, <marker 0>
<time 1>, <load 1>, <PV 1>, <price 1>, <marker 1>
...

Decision requirements:
- Keep adequate reserves for later expensive periods; pre-charge when prices are low or PV is abundant.
- Avoid wasting PV (curtailment) when storage headroom exists.
- Prevent deep depletion before long high-load stretches.

Return format:
- Output exactly one of [0], [1], [2], [3], [4].
\end{lstlisting}

This consistent format is important for the transfer test in Section~\ref{subsec:battery-transfer}. It lets the battery experiment probe whether TES-trained reasoning patterns remain useful when the physical semantics change, without simultaneously changing the interaction format.

\subsection{Forecast-error prompt augmentation}
Under forecast-error conditions, the TES prompt is augmented with the following additional user-message block. This is the explicit wording used to make forecast uncertainty visible to the LLM without changing the action interface itself:
\begin{lstlisting}[style=promptblock]
Forecast data:
<forecast table>

Forecast caveat:
- The forecast for cooling demand is not 100% accurate, and errors may occur.
- Please analyze the accuracy and trends of the prediction model based on the errors observed so far.
- Keep decisions within the TES operating bounds, leaving buffer for possible underestimation.

Operational notes:
...
- Forecasts can underestimate or overestimate demand, so leave safety margin when near the bounds.
\end{lstlisting}

\end{document}